\documentclass[10pt,twocolumn,letterpaper]{article}

\usepackage{cvpr}              
\usepackage{bm}
\usepackage[accsupp]{axessibility}

\usepackage{colortbl}
\definecolor{rankfirst}{HTML}{FFCCCC}
\definecolor{ranksecond}{HTML}{CCE5FF}
\definecolor{rankthird}{HTML}{D6F5D6}
\newcommand{\first}[1]{{\setlength{\fboxsep}{0.5pt}\colorbox{rankfirst}{\textbf{#1}}}}
\newcommand{\second}[1]{{\setlength{\fboxsep}{0.5pt}\colorbox{ranksecond}{#1}}}
\newcommand{\third}[1]{{\setlength{\fboxsep}{0.5pt}\colorbox{rankthird}{#1}}}

\definecolor{cvprblue}{rgb}{0.21,0.49,0.74}
\usepackage[pagebackref,breaklinks,colorlinks,allcolors=cvprblue]{hyperref}

\def\paperID{xxxx} 
\def\confName{CVPR}
\def\confYear{2026}

\title{RiGS: Rigid-aware 4D Gaussian Splatting from a Single Monocular Video}

\author{
 Chenyu Wu$^{1,2,}$\thanks{Work done during a research internship at Harvard University.} \quad Wanhua Li$^{1,3,}$\thanks{Corresponding author.} \quad Zhu-Tian Chen$^{4}$ \quad Hanspeter Pfister$^{1}$ \\
 $^{1}$Harvard University \quad $^{2}$Zhejiang University \\ $^{3}$Nanyang Technological University \quad $^{4}$University of Minnesota - Twin Cities  \\
 {\tt\small chenyuwu@zju.edu.cn, wanhua.li@ntu.edu.sg, ztchen@umn.edu, pfister@seas.harvard.edu}
}

\begin{document}
\maketitle
\begin{abstract}
Reconstructing dynamic 3D scenes from monocular videos is a fundamental yet highly challenging task, as real-world motions often involve both long-term smooth transformations and short-term complex deformations. Existing methods either struggle to maintain temporal consistency or fail to capture high-frequency dynamics due to limited motion modeling capacity. In this work, we present Rigid-aware 4D Gaussian Splatting (RiGS), which simultaneously captures motions across multiple temporal scales. 
Specifically, RiGS introduces three types of Gaussian primitives: static, rigid, and transient, which represent static backgrounds, long-term low-frequency motions, and short-term high-frequency dynamics, respectively.
An object-wise dynamic mask is proposed to aggregate long-range spatiotemporal motion information and guide the decomposition of static and dynamic regions. To jointly model motion across scales, rigid Gaussians are allowed to transition into transient Gaussians based on their temporal duration, and both are optimized under scene flow guidance, providing dense 3D motion supervision. Extensive experiments demonstrate that RiGS achieves state-of-the-art performance on novel view synthesis benchmarks. Code is available at \hyperlink{https://github.com/ladvu/RiGS}{https://github.com/ladvu/RiGS}.
\end{abstract}    
\section{Introduction}
\label{sec:intro}

Dynamic view synthesis~\cite{luo2025instant4d,wu2025orientationanchoredhypergaussian4dreconstruction,chen2025prodygprogressivedynamicscene} from monocular videos is a cornerstone problem in computer vision and computer graphics, with broad applications in virtual and augmented reality (VR/AR)~\cite{wu20244dgaussiansplattingrealtime}, autonomous driving~\cite{peng2025desire}, and robotics~\cite{luiten2024dynamic}. Accurate reconstruction of dynamic 3D scenes enables immersive scene exploration, motion understanding, and safe interaction in real-world environments. Among various sensing modalities, monocular video is one of the most ubiquitous and accessible sources of visual data, inherently rich in dynamic motion. Consequently, reconstructing dynamic 3D scenes from a single video remains a fundamental yet highly challenging task of both academic and practical significance.


Several recent approaches~\cite{wang2025gflow, liu2023robust, stearns2024marbles} perform per-frame optimization, which can capture high-frequency motion details but often suffer from poor temporal consistency and unstable geometry across frames. More recent works~\cite{som2024, Lei_mosca_2025_CVPR, guo2025uncertaintymattersdynamicgaussian} adopt the as-rigid-as-possible assumption, which constrains local motion using rigid transformations to maintain geometric stability. Such methods perform well for objects exhibiting coherent and low-frequency motion and are more robust under large viewpoint variations. 
However, the limited motion bases and rigid constraints make these methods less effective in modeling regions with large deformation or fast, non-rigid motion, often leading to incomplete reconstructions.

\begin{figure}[t]
    \centering
    \includegraphics[width=1\linewidth]{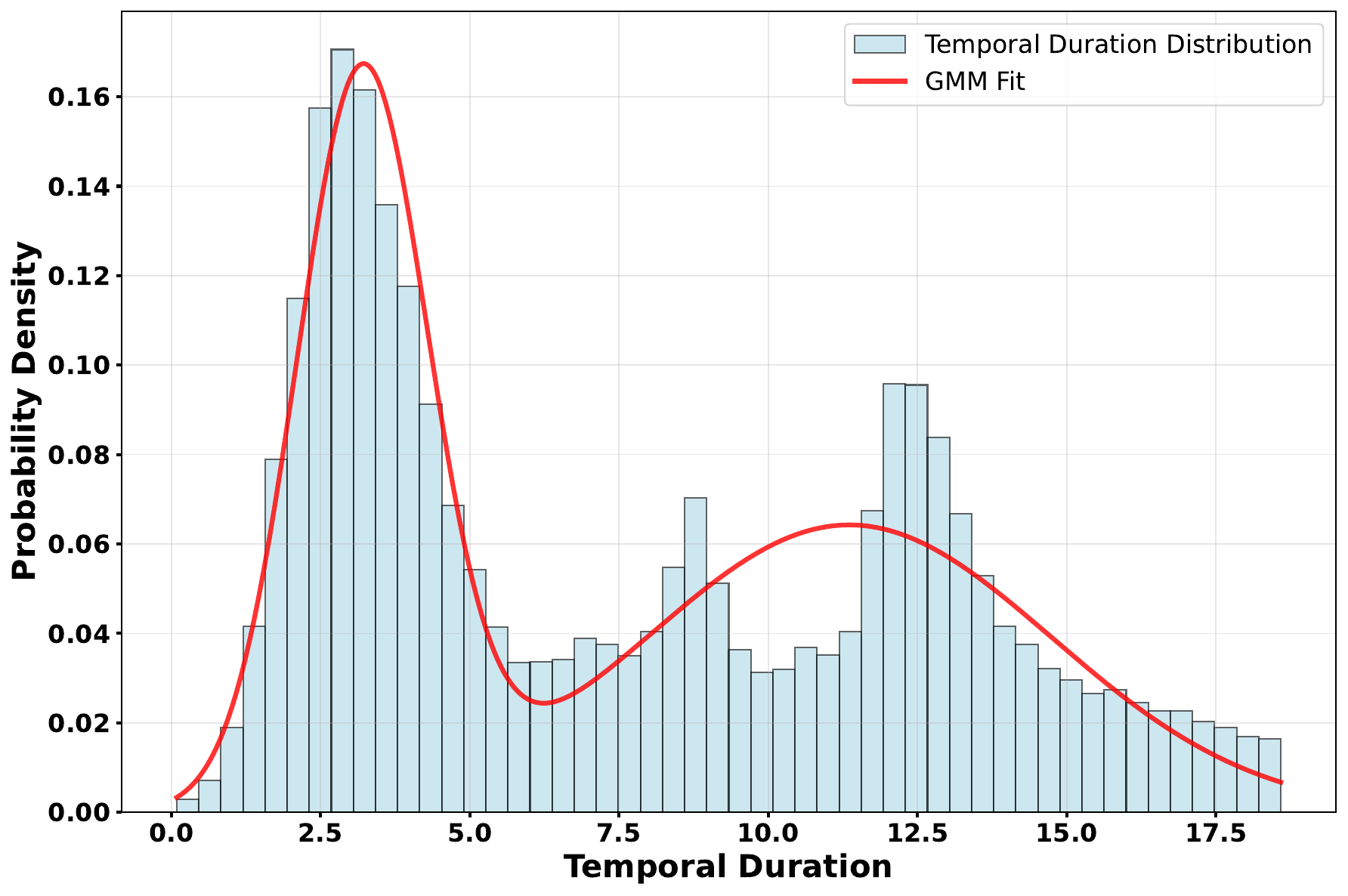}
    \caption{Distribution of temporal durations (in frame indices) of rigid-based dynamic Gaussians, computed from the \textit{Umbrella} scene in the Nvidia dynamic scene dataset. The two peaks indicate that rigid-based dynamic Gaussian representations capture long-term smooth motion but struggle with short-term complex dynamics. We observe this two-peak pattern consistently across various scenes (see supplementary material).}
    \label{fig:distribution}
    \vspace{-10pt}
\end{figure}

To further investigate this limitation, we analyze dynamic scenes using a rigid-based Gaussian representation with temporal opacity decay, which allows each Gaussian to exist only for a limited time within the video.
After training, we measure the temporal duration of all dynamic Gaussians, and the resulting distribution is shown in Figure~\ref{fig:distribution}.
Interestingly, the distribution exhibits two distinct peaks: one at large temporal durations, corresponding to Gaussians modeling long-term, low-frequency rigid motion, and another sharp peak at short durations, where Gaussians attempt to fit short-term, high-frequency complex motion.
This bimodal pattern reveals that while rigid-based representations effectively capture stable motion, they struggle to represent short-term, fine-grained dynamics, which highlights the fundamental challenge of modeling motion across multiple temporal scales.



To address these challenges, we propose Rigid-aware 4D Gaussian Splatting (RiGS), a representation that models dynamic scenes across multiple temporal scales.
RiGS adaptively decomposes scene motion into three types of Gaussian primitives: static Gaussians that represent time-invariant backgrounds, rigid Gaussians that capture long-term, low-frequency motions, and transient Gaussians that handle short-term, high-frequency dynamic effects.
Rigid Gaussians are designed to model coherent and stable motion through $\mathbb{SE}(3)$ transformations, while transient Gaussians are dynamically activated to capture fast or non-rigid movements with greater flexibility.
By jointly optimizing these components under scene flow guidance, RiGS achieves temporally consistent, fine-grained, and geometrically accurate reconstruction of dynamic 3D scenes.

In summary, our core contributions are threefold:
\begin{itemize}
    \item We propose RiGS, which jointly models static, rigid, and transient Gaussians, enabling multi-scale motion modeling within a unified framework.
    \item We introduce an object-wise dynamic mask for static–dynamic decomposition and a joint optimization strategy guided by dense 3D scene flow for learning motion across multiple scales.
    \item Our method achieves the state-of-the-art performance on several challenging monocular dynamic scene benchmarks, showing the superiority of our proposed method.
\end{itemize}

\section{Related Works}
\label{sec::related_works}
\textbf{Novel View Synthesis on Static Scenes.}
Novel view synthesis (NVS) is the task of generating images of a scene from new, unseen viewpoints, given a set of images captured from known viewpoints. A significant breakthrough came with the introduction of Neural Radiance Fields (NeRF)~\cite{mildenhall2020nerf, barron2021mipnerf}, which represent a scene implicitly with a multi-layer perceptron (MLP) that maps a 3D coordinate and viewing direction to volume density and view-dependent color. However, its sampling-based volumetric rendering leads to prohibitively long training times.
Subsequent works~\cite{mueller2022instant, yu2021plenoxelsradiancefieldsneural} employ explicit voxel grids of feature vectors to enable faster optimization.
More recently, 3D Gaussian Splatting (3DGS)~\cite{kerbl3Dgaussians, Huang2DGS2024,fan2025momentum} has emerged as a compelling explicit alternative. It models a scene as a collection of anisotropic 3D Gaussians—primitives with optimizable positions, covariances, opacities, and spherical harmonic coefficients. 
3DGS has rapidly gained widespread adoption across a variety of domains~\cite{qin2024langsplat,lilangsplatv2,chen2024text,kocabas2024hugs,han2025reparo} due to its superior performance.

\noindent\textbf{Novel View Synthesis on Dynamic Scenes.} 
Many methods~\cite{bansal20204dvisualizationdynamicevents, pumarola2020dnerf, wang2025freetimegs} have been proposed for multi-view video inputs.
For example, Zhang \emph{et al.}~\cite{zhang2025occnerf} propose a novel framework named OccNeRF, which effectively predicts 3D Occupancy from multi-view inputs in a self-supervised manner and eliminates the reliance on the expensive LiDAR point clouds.
For monocular videos, previous approaches~\cite{liu2023robust, du2021nerflow, Fang_2022, Gao-ICCV-DynNeRF} adopt NeRF and introduce additional parameters to model spatiotemporal variations.
Others follow a template-based paradigm~\cite{park2021hypernerf, park2021nerfies}, learning a canonical space and a deformation field that maps observed frames to this space.
While these implicit methods can produce high-quality results, they are often restricted to a limited range of viewpoints and suffer from the inherently slow training of NeRF.
Recent methods employ Gaussian Splatting for monocular dynamic scene reconstruction.
Several works~\cite{yang2023deformable3dgs, wu20244dgaussiansplattingrealtime, luo2025instant4d, stearns2024marbles} model motion by predicting per-Gaussian positional offsets, offering flexibility but sometimes struggling with temporal coherence.
Another line of work~\cite{som2024, Lei_mosca_2025_CVPR, wu2025orientationanchoredhypergaussian4dreconstruction, guo2025uncertaintymattersdynamicgaussian, chen2025prodygprogressivedynamicscene} learns explicit motion trajectories to guide the movement of Gaussians over time, which can better capture long-term dynamics but may lack expressiveness for high-frequency changes. 
4D-Fly~\cite{Wu_2025_CVPR} and Instant4D~\cite{luo2025instant4d} focus on accelerating the optimization process, achieving fast 4D reconstruction from monocular videos.
BTimer~\cite{liang2025feedforwardbullettimereconstructiondynamic} and MoVieS~\cite{lin2026movies} adopt feed-forward prediction paradigms that bypass per-scene optimization entirely.
4D3R~\cite{guo20254d3rmotionawareneuralreconstruction} investigates 4D reconstruction from monocular videos with constraint input.
While these existing methods struggle to effectively represent motion across multiple temporal scales, we aim to address this issue through the proposed Rigid-aware 4D Gaussian Splatting. 
\section{Method}
\subsection{Preliminaries}

3D Gaussian Splatting~\cite{kerbl3Dgaussians} has emerged as an expressive and efficient representation for 3D scene modeling and rendering, which we adopt as the basis of our method. A 3D Gaussian primitive $g_i$ is parameterized by mean $\bm{\mu}_i \in \mathbb{R}^3$, scale $\mathbf{s}_i \in \mathbb{R}^3$,  rotation quaternion $\mathbf{q}_i \in \mathbb{R}^4$, opacity $o_i \in \mathbb{R}$, and color $\mathbf{c}_i \in \mathbb{R}^3$. The 3D covariance matrix is formulated as $\mathbf{\Sigma}_i = \mathbf{R}_i\mathbf{S}_i\mathbf{S}_i^T\mathbf{R}_i^\mathbf{T}$, where $\mathbf{R}_i \in \mathbb{R}^{3\times3}$ and $\mathbf{S}_i \in \mathbb{R}^{3\times 3} $ are the rotation and scale matrix obtained from $\mathbf{q}_i$ and $\mathbf{s}_i$, respectively. To enable efficient and differentiable rendering, the 2D covariance matrix of $g_i$ is approximated by an affine transformation:
\begin{equation}
  \mathbf{\Sigma'} = \mathbf{J} \mathbf{W} \mathbf{\Sigma} \mathbf{W}^T \mathbf{J}^T,
  \label{eq:ewa}
\end{equation}
where $\mathbf{W}$ is the viewing transformation and $\mathbf{J}$ is the affine approximation of the projective transformation~\cite{zwicker2001ewa, zwicker2001surface}. Let the projected 2D mean of a Gaussian $g_i$ be $\bm{\mu}_i'$. Then the final output color at pixel $\mathbf{p}$ is calculated by alpha blending:
\begin{equation}
    \mathbf{C}(\mathbf{p}) = \sum_{i \in N} \mathbf{c}_i  \alpha_i  \prod_{j=1}^{i-1}(1 - \alpha_j),
    \label{eq:alpha_blending}
\end{equation}
\begin{equation}
    \alpha_i = o_i \cdot \exp(- \frac{1}{2}(\mathbf{p} - \bm{\mu}_i')^T \mathbf{\Sigma}'^{-1}(\mathbf{p} - \bm{\mu}_i')).
\end{equation}

\begin{figure*}[t]
    \centering
    \includegraphics[width=1.0\linewidth]{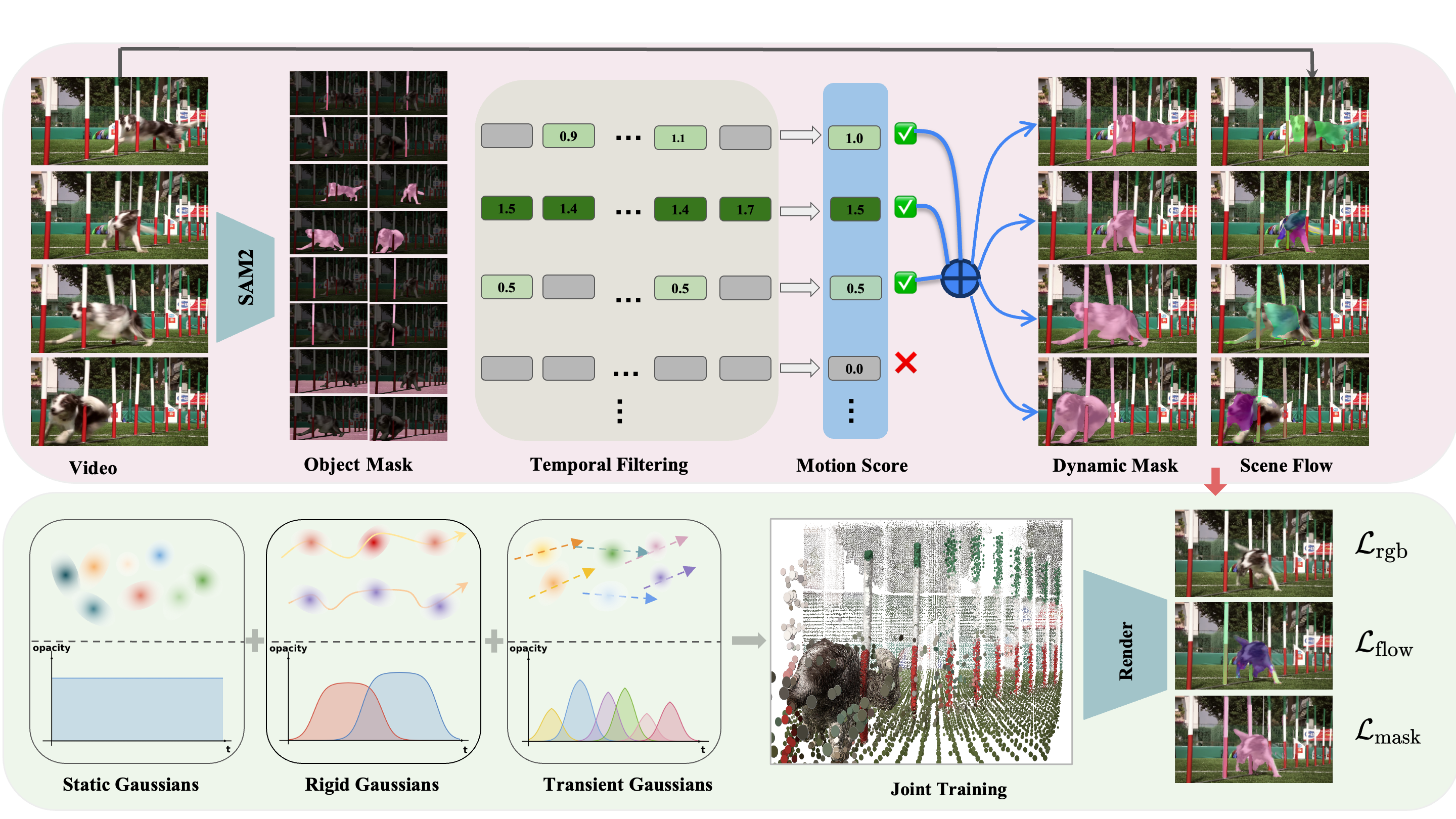}
    \caption{The pipeline of our proposed RiGS. RiGS consists of three types of Gaussian primitives: static, rigid, and transient. We propose object-wise dynamic masks to supervise static–dynamic decomposition. Rigid Gaussians capture long-term, low-frequency motions, and transient Gaussians model short-term, high-frequency dynamics. These dynamic Gaussians are jointly trained with scene flow guidance. }
    \label{fig:Method}
    \vspace{-12pt}
\end{figure*}

While 3DGS is primarily designed for reconstructing static scenes, extending it to dynamic videos requires modeling the temporal dimension of the Gaussian points.
Several studies have extended 3D Gaussian Splatting to support 4D scene reconstruction. 4DGS~\cite{wu20244dgaussiansplattingrealtime} employs MLPs to predict the deformation of Gaussian positions and shapes over time. Similar work~\cite{yang2023gs4d} generalizes 3D Gaussians into 4D primitives, where the temporal deformations are encoded by a 4D covariance matrix.


\subsection{Method Overview}
Reconstructing 4D scene using only a single monocular video is a challenging problem.
Existing methods struggle to effectively represent motions across multiple temporal scales. To address this issue, we propose \textbf{RiGS}, a rigid-aware 4D reconstruction framework that simultaneously captures motions across multiple temporal scales.
Specifically, RiGS introduces three types of Gaussian primitives: static Gaussians $\{g^s\}$, rigid Gaussians $\{g^r\}$, and transient Gaussians $\{g^t\}$. Static Gaussians $g^s = (\bm{\mu}^s, \mathbf{s}^s, \mathbf{q}^s, o^s, \mathbf{c}^s)$ represent time-invariant components that typically correspond to the static background and persist throughout the entire video. 
To capture dynamic content, Rigid Gaussians $g^r$ are introduced to model long-term, low-frequency smooth motions, while transient Gaussians $g^t$ are designed to represent short-term, high-frequency complex motions.

Given an image sequence from a single monocular video $\mathcal{I} = \{\mathbf{I}_i\}_{i=1}^T$, where $\mathbf{I}_i \in \mathbb{R}^{H\times W}$ denotes the $i$-th frame, we first leverage priors from powerful 2D foundation models and video pose engine~\cite{huang2025vipe, li2025megasam} to obtain additional frame-wise supervisions. Specifically, for the frame at time $t$, we obtain camera intrinsics $\mathbf{K}_t$, extrinsics $\mathbf{E}_t$, metric depth $\mathbf{D}_t$, forward optical flow $\mathbf{F}^{\text{fwd}}_t$, backward optical flow $\mathbf{F}^{\text{bwd}}_t$, and 2D tracking points $\{\mathbf{U}_{t \to i}\}_{i=1}^T$.
To better guide the dynamic–static decomposition, we propose object-wise dynamic masks that aggregate long-range, object-level spatiotemporal motion information. This provides accurate and efficient supervision for learning both dynamic and static Gaussian points (Sec.~\ref{sec:decomposition}).
To effectively model motions at different temporal scales, we jointly learn rigid Gaussians and transient Gaussians for dynamic regions.
Rigid Gaussians can adaptively transform into transient Gaussians depending on their temporal duration during optimization.
Furthermore, we employ scene flow to guide the 3D motion directions of both rigid and transient Gaussians, facilitating effective multi-scale motion modeling (Sec.~\ref{sec:dynamic_scene}). Figure~\ref{fig:Method} visualizes the overall pipeline of our proposed method.

\subsection{Static Dynamic Decomposition}
\label{sec:decomposition}
Static–dynamic decomposition has proven to be an essential step in monocular dynamic scene reconstruction~\cite{Lei_mosca_2025_CVPR,som2024, liu2023robust}.
During rendering, static Gaussians are constrained to reconstruct areas outside the dynamic mask, whereas dynamic Gaussians are responsible for regions within the mask. This supervision effectively guides the decomposition of Gaussians into static and dynamic components and contributes to both improved reconstruction quality and reduced computational cost.
Previous methods~\cite{Lei_mosca_2025_CVPR, liu2023robust, huang2025vipe, rockwell2025dynamiccameraposes, golisabour2024romo, Huang_2025_CVPR_seganymo, chen2025easi3r, luo2025instant4d} primarily focus on pixel-level dynamic mask prediction, either by estimating motion cues (e.g., optical flow) or leveraging semantic segmentation priors. Such approaches inevitably treat each pixel independently and fail to maintain temporal and spatial coherence across entire objects.
 

We make two key observations about motion in real-world videos:
\textbf{1)} The motion of an object may not persist throughout the video—for instance, a car remains still at a red light and moves only when it turns green;
\textbf{2)} The motion may occur only on a local part of an object—for example, a dog wagging its tail while keeping the rest of its body stationary.
Pixel-level prediction tends to segment these locally static or dynamic parts separately across time and space, causing a single object to be inconsistently represented by both static and dynamic Gaussians.
This leads to inevitable discontinuities and inconsistencies in the overall modeling.
We argue that if any part of an object exhibits motion at any time, the entire object should be treated as dynamic.
This prevents static and dynamic Gaussians from redundantly modeling different spatiotemporal portions of the same object and ensures a globally consistent representation.
To this end, we propose object-wise dynamic masks, which determine dynamic regions at the object level by aggregating spatiotemporal motion information across the entire video (see Figure~\ref{fig:Method}).

Specifically, we first extract individual objects and their corresponding IDs using SAM2~\cite{ravi2024sam2segmentimages}.
We then compute a motion score for each object to quantify its dynamic degree over time.
Let $\mathcal{P}_{i,t}$ denote the set of pixels belonging to object $i$ in frame $t$, and let $s_{i,t}$ represent its frame-wise motion score, defined as:
\begin{equation}
    s_{i,t} = \frac{1}{\sum_{\mathbf{p} \in \mathcal{P}_{i,t}}w_{t}(\mathbf{p})} \sum_{\mathbf{p} \in \mathcal{P}_{i,t}} w_t({\mathbf{p}) e_t(\mathbf{p})},
    \label{eq:dynmask}
\end{equation}
where $w_t(\mathbf{p})$ is a weighting term derived from flow uncertainty~\cite{raft} and flow mask~\cite{meister2017unflow}, and $e_t(\mathbf{p})$ is the Sampson error~\cite{sampson} at frame $t$.
We then apply temporal filtering using a threshold $\epsilon^{\text{temp}}$ to identify frames in which object $i$ exhibits significant motion: $\mathcal{F}_i = \{t \mid s_{i,t} > \epsilon^{\text{temp}}\}$. 
The overall motion score $s_i$ of object $i$ is computed by aggregating its motion scores across these motion frames:
\begin{equation}
    s_{i} = \frac{1}{|\mathcal{F}_i|} \sum_{t \in \mathcal{F}_i} s_{i,t}.
    \label{eq:score}
\end{equation}
Finally, the dynamic mask $\mathbf{M}^{\text{dyn}}_t$ is obtained by combining all object masks 
$\mathbf{M}^{\text{obj}}_{i,t}$ whose motion scores exceed the dynamic threshold $\epsilon^{\text{dyn}}$: $\mathbf{M}^{\text{dyn}}_t = 
    \bigcup_{i:\, s_i > \epsilon^{\text{dyn}}} \mathbf{M}^{\text{obj}}_{i,t}.$

After obtaining the object-wise dynamic masks, we use them to supervise the training of static Gaussians $\{g^s\}$ and dynamic Gaussians $\{g^r,g^t\}$.
Specifically, we render a predicted dynamic mask $\hat{\mathbf{M}}$ by adding one additional feature channel and assigning a value of 0 to static Gaussians and 1 to dynamic Gaussians.
The rendered mask $\hat{\mathbf{M}}$ is then compared with the ground-truth $\mathbf{M}^{\text{dyn}}$ using binary cross-entropy loss:
\begin{equation}
\mathcal{L}_{\text{mask}} = \text{BCE}(\hat{\mathbf{M}}, \mathbf{M}^{\text{dyn}}).
\label{eq:loss_mask}
\end{equation}

This mask supervision effectively guides the decomposition between static and dynamic Gaussians, ensuring that dynamic regions are accurately captured while avoiding the leakage of motion into static components.

\subsection{Rigid-aware Dynamic Gaussians} 
\label{sec:dynamic_scene}
To model motion dynamics across multiple temporal scales, we employ three sets of Gaussian primitives.
The static Gaussians $\{g^s\}$ are initialized from the segmentation defined by $\mathbf{M}^{\text{dyn}}$ and represent the static background.
Within dynamic regions, we introduce rigid Gaussians $\{g^r\}$ and transient Gaussians $\{g^t\}$ to capture long-term consistent and short-term high-frequency motions, respectively.

\noindent\textbf{Rigid Gaussians.} 
To effectively capture long-term, low-frequency motion, we design rigid Gaussians that explicitly incorporate transformations in $\mathbb{SE}(3)$, representing rotation and translation without deformation. Such transformations naturally preserve spatial relationships, making them well-suited for modeling stable, rigid motion over extended time periods. Motivated by prior work~\cite{som2024}, we parameterize each rigid Gaussian as
$g^r = (\bm{\mu}^r, \mathbf{s}^r, \mathbf{q}^r, o^r, \mathbf{c}^r, \mathbf{w},\beta^r, \gamma^r)$ and define motion bases $\{\{\mathbf{T}_{j,t}\in \mathbb{SE}(3)\}_{j=1}^K\}_{t=1}^T$,
where $\mathbf{T}_{j,t}$ denotes the $j$-th motion basis at time $t$, and $\mathbf{w}$ is the weighting vector controlling their combination.
At a query time $t$, the rigid Gaussian is transformed from its canonical position and rotation $(\bm{\mu}^r, \mathbf{R}^r)$ as follows:\
\begin{equation}
    \bm{\mu}^r(t) = (\sum_{j=1}^{K}\mathbf{w}_{j} \mathbf{T}_{j,t})\bm{\mu}^{r},
    \label{eq:som_transform_mean}
\end{equation}
\begin{equation}
    \mathbf{R}^r(t) = (\sum_{j=1}^{K}\mathbf{w}_{j} \mathbf{T}_{j,t})_{:3,:3} \mathbf{R}^{r},
    \label{eq:som_transform_r}
\end{equation}
where $\|\mathbf{w}\|_2^2=1$ ensures normalized motion weighting. During motion trajectory modeling, we linearly interpolate the 6D continuous rotation representation~\cite{zhou2020continuityrotationrepresentationsneural} alongside the 3D translation, and then convert the result to a valid rotation matrix.

This formulation enforces a low-rank motion constraint, allowing groups of rigid Gaussians to move coherently under shared transformations.
It excels at capturing stable, long-term motions of rigid objects.
However, because the number of motion bases $K$ is fixed during training, this formulation lacks flexibility in modeling short-term, high-frequency, or locally complex motions, limiting its expressiveness for rapidly deforming or fine-grained dynamic regions.
To tackle the above problems, we introduce a soft gating function to model the temporal opacity decay of each rigid Gaussian:
\begin{equation}
    o(t) = o \sigma(\alpha(\beta - |t-\gamma|)),
    \label{eq::sigmoid_decay}
\end{equation}
where $\alpha \in \mathbb{R}^+$ controls the boundary sharpness and $\sigma$ is the sigmoid function. $\beta \in \mathbb{R}^+$ denotes the \textbf{temporal duration} over which a rigid Gaussian remains active, and $\gamma \in \mathbb{R}$ is its temporal center. This allows each rigid Gaussian to exist only within a limited time range of the video.
$\beta$ and $\gamma$ are initialized from tracking visibilities, while the remaining parameters are initialized following~\cite{som2024}.
A regularization term on $\beta^r$ (Eq.~\ref{eq::reg_loss}) is further applied to prevent rigid Gaussians from degenerating into frame-specific overfitting.

\noindent\textbf{Transient Gaussians.} 
By training with rigid Gaussians using temporal opacity decay, we observe a distinct two-peak pattern in the distribution of $\beta^r$, as shown in Figure~\ref{fig:distribution}. Rendering results reveal that rigid Gaussians with large $\beta^r$ tend to fit the main body of dynamic objects exhibiting smooth, low-frequency motion, whereas those with small $\beta^r$ correspond to regions with rapid or complex motion. This bimodal distribution suggests that the low-rank motion bases in rigid Gaussians are insufficient to model short-term, high-frequency dynamics. To overcome this limitation, we introduce transient Gaussians to explicitly model fast and complex short-term motion in dynamic regions. We define each transient Gaussian as
$g^t = (\bm{\mu}^t, \mathbf{s}^t, \mathbf{q}^t, o^t, \mathbf{c}^t, \mathbf{v},\beta^t, \gamma^t)$, where $\mathbf{v}$ denotes the learnable motion velocity vector. At a specific timestamp $t$, its mean position is computed as:
\begin{equation}
    \bm{\mu}^t(t) = \bm{\mu}^t + \mathbf{v} (t - \gamma^t)
    \label{eq:linear}
\end{equation}
and the opacity $o^t$ follows the function defined in Eq.~\ref{eq::sigmoid_decay}.

When $\beta^t$ is small, each transient Gaussian is active only within a short temporal window, during which its motion approximates a linear trajectory.
Unlike rigid Gaussians constrained by shared motion bases, transient Gaussians have fully learnable motion directions $\mathbf{v}$, allowing them to move freely in any direction.
By aggregating multiple short-duration transient Gaussians, our model can effectively capture non-rigid, fine-grained, and high-frequency motion patterns that rigid Gaussians alone cannot represent.

\begin{table*}[t]
  \caption{Quantitative Evaluation on Nvidia Dynamic Scene Dataset}
  \label{tab:Nvidia}
  \centering
  \setlength{\tabcolsep}{2pt}
  \begin{tabular}{@{}lcccccccc@{}}
    \toprule
    PSNR$\uparrow$/LPIPS$\downarrow$ & Jumping & Skating & Truck & Umbrella & Balloon1 & Balloon2 & Playground  & AVE\\
    \midrule
    D-NeRF~\cite{pumarola2020dnerf} & 22.36/0.193 & 22.48/0.323 & 24.10/0.145 & 21.47/0.264 & 19.06/0.259 & 20.76/0.277 & 20.18/0.164 & 21.49/0.232 \\
    TiNeuVox~\cite{TiNeuVox} & 20.81/0.247 & 23.32/0.152 & 23.86/0.173 & 20.00/0.355 & 17.30/0.353 & 19.06/0.279 & 13.84/0.437 & 19.74/0.285 \\
    DynNeRF~\cite{Gao-ICCV-DynNeRF} & 24.68/\third{0.090} & \third{32.66}/\third{0.035} & \third{28.56}/0.082 & 23.26/0.137 & 22.36/0.104 & \third{27.06}/\second{0.049} & 24.15/0.080 & 26.10/0.082 \\
    RoDynRF~\cite{liu2023robust} & \second{25.66}/\first{0.071} & 28.68/0.040 & \second{29.13}/\third{0.063} & 24.26/\third{0.089} & 22.37/0.103 & 26.19/0.054 & \second{24.96}/\third{0.048} & 25.89/\second{0.067} \\
    4DGS~\cite{wu20244dgaussiansplattingrealtime} & 21.93/0.269 & 24.84/0.174 & 23.02/0.175 & 21.83/0.213 & 21.32/0.185 & 18.81/0.178 & 18.40/0.196 & 21.45/0.199 \\
    DynPoint~\cite{zhou2025dynpointdynamicneuralpoint} & 24.69/0.097 & 31.34/0.045 & \first{29.30}/\second{0.061} & \third{24.59}/\second{0.086} & \third{22.77}/\second{0.099} & \second{27.63}/\second{0.049} & \first{25.37}/\first{0.039} & \third{26.53}/\third{0.068} \\
    MoSca~\cite{Lei_mosca_2025_CVPR} & \third{25.01}/\third{0.090} & \second{33.41}/\second{0.030} & 27.83/0.080 & \second{25.17}/0.090 & \second{23.58}/\third{0.100} & 27.80/\third{0.050} & 24.25/0.050 & \second{26.72}/0.070 \\
    Ours & \first{25.78}/\second{0.072} & \first{34.79}/\first{0.021} & 28.40/\first{0.047} & \first{25.84}/\first{0.058} & \first{24.24}/\first{0.083} & \first{28.25}/\first{0.032} & \third{24.73}/\second{0.043} & \first{27.43}/\first{0.051} \\
    \bottomrule
  \end{tabular}
\end{table*}

\begin{figure*}[t]
  \centering
  \label{fig:nvidia_qualitative}
  \begin{subfigure}[t]{\textwidth}
    \centering
    \includegraphics[width=0.16\textwidth]{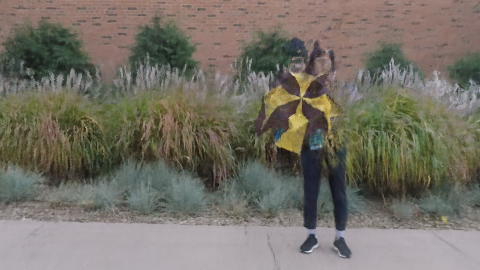}
    \includegraphics[width=0.16\textwidth]{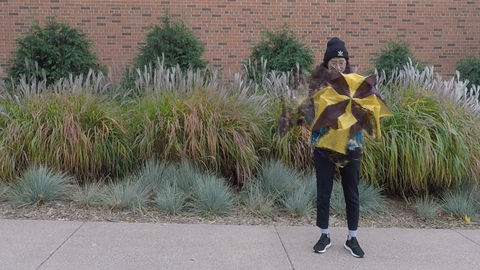}
    \includegraphics[width=0.16\textwidth]{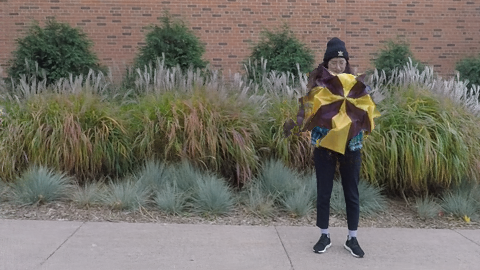}
    \includegraphics[width=0.16\textwidth]{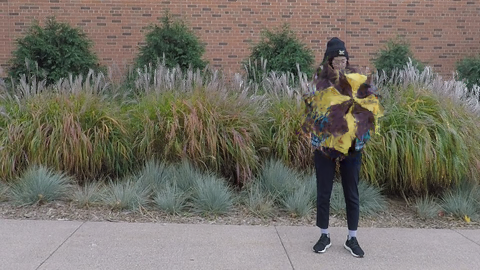}
    \includegraphics[width=0.16\textwidth]{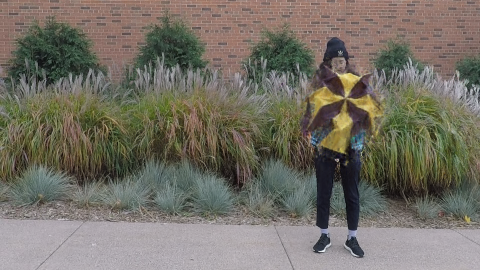}
    \includegraphics[width=0.16\textwidth]{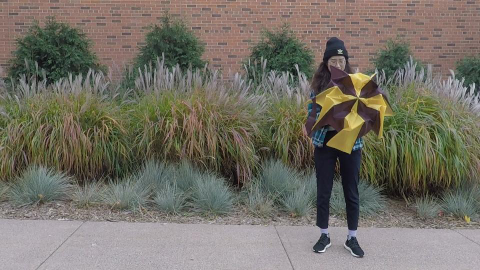}
  \end{subfigure}

  \vspace{4pt}

  \begin{subfigure}[t]{\textwidth}
    \centering
    
    \begin{subfigure}[t]{0.16\textwidth}
      \centering
      \includegraphics[width=\linewidth]{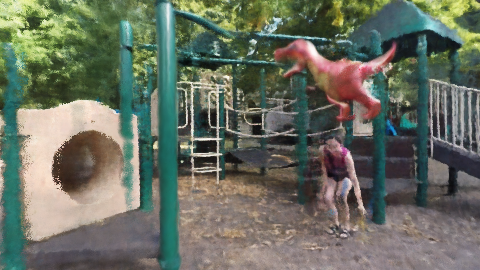}
      \caption*{\footnotesize HyperNeRF~\cite{park2021hypernerf}}
    \end{subfigure}
    \begin{subfigure}[t]{0.16\textwidth}
      \centering
      \includegraphics[width=\linewidth]{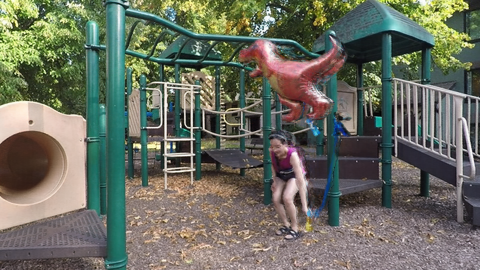}
      \caption*{\footnotesize RoDynRF~\cite{liu2023robust}}
    \end{subfigure}
    \begin{subfigure}[t]{0.16\textwidth}
      \centering
      \includegraphics[width=\linewidth]{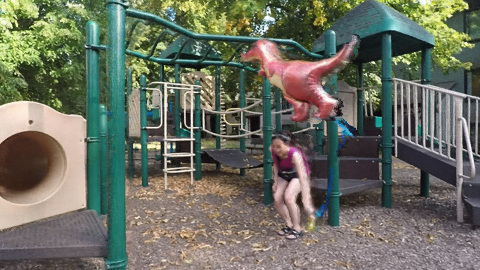}
      \caption*{\footnotesize DynNeRF~\cite{Gao-ICCV-DynNeRF}}
    \end{subfigure}
    \begin{subfigure}[t]{0.16\textwidth}
      \centering
      \includegraphics[width=\linewidth]{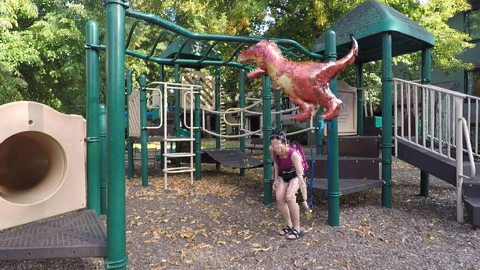}
      \caption*{\footnotesize MoSca~\cite{Lei_mosca_2025_CVPR}}
    \end{subfigure}
    \begin{subfigure}[t]{0.16\textwidth}
      \centering
      \includegraphics[width=\linewidth]{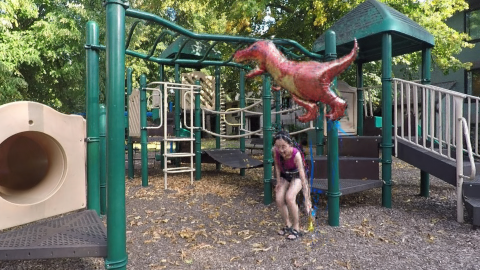}
      \caption*{\footnotesize Ours}
    \end{subfigure}
    \begin{subfigure}[t]{0.16\textwidth}
      \centering
      \includegraphics[width=\linewidth]{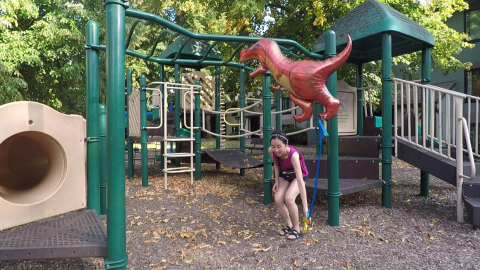}
      \caption*{\footnotesize Ground Truth}
    \end{subfigure}
  \end{subfigure}
  \caption{
    \textbf{Evaluation on Nvidia Dynamic Scenes dataset.} Our method achieves state-of-the-art performance in novel view synthesis. We present qualitative comparisons on the Umbrella and Playground scenes, which respectively contain large non-rigid deformations and complex multi-object motions. NeRF-based approaches struggle with these challenges and often produce inconsistent geometry. The previous SOTA method~\cite{Lei_mosca_2025_CVPR} also fails to handle the significant deformation in the Umbrella scene, leading to noticeable artifacts. In contrast, our method accurately models the umbrella’s deformation and preserves finer details in the Playground scene.
  }
  \label{fig:qualitative_nvidia}
  \vspace{-10pt}
\end{figure*}

\noindent\textbf{Joint Optimization.}   
We begin training with a warm-up stage, where static Gaussians, rigid Gaussians, and motion bases are first initialized and optimized to establish stable geometry and coarse motion patterns.
After warm-up, rigid Gaussians with short temporal duration $\beta^r$ are converted into transient Gaussians to better capture fine-grained short-term dynamics. Subsequently, all three types of Gaussians—static $\{g^s\}$, rigid $\{g^r\}$, and transient $\{g^t\}$—are jointly optimized together with the motion bases.

During joint training, any rigid Gaussian whose temporal duration $\beta^r$ becomes smaller than the threshold is dynamically transformed into a transient Gaussian.

\noindent\textbf{Scene Flow Guidance.} 
For transient Gaussians, the supervision signals are often sparse, and their motion parameters possess high degrees of freedom, making effective learning challenging. To better guide the motion learning of dynamic Gaussians, we propose to use scene flow as a dense and geometry-aware supervision signal.

We first lift the optical flow to a 3D scene flow using camera parameters and metric depth. Let the forward and backward scene flow of frame $t$ be $\mathbf{v}^{\text{fwd}}_t$ and $\mathbf{v}^{\text{bwd}}_t$, respectively:
\begin{equation}
    \mathbf{v}_t^{\text{fwd}} =
    \mathcal{W}\big(\pi^{-1}_{t+1}(\mathbf{D}_{t+1}), \mathbf{F}^{\text{fwd}}_t\big)
    - \pi^{-1}_t(\mathbf{D}_t),
    \label{eq:fwd_v}
\end{equation}
\begin{equation}
    \mathbf{v}_t^{\text{bwd}} = \pi^{-1}_t(\mathbf{D}_t) - \mathcal{W}(\pi^{-1}_{t-1}(\mathbf{D}_{t-1}), \mathbf{F}^{\text{bwd}}_t),
    \label{eq:bwd_v}
\end{equation}
where $\mathcal{W}$ is the warping operator and $\pi$ is the projection operator parameterized by $\mathbf{E}$ and $\mathbf{K}$. We further compute the scene flow mask as the intersections of the dynamic mask, depth mask, warped depth mask, and flow mask~\cite{meister2017unflow}. 
During training, we render the forward and backward velocities of the Gaussians, $\hat{\mathbf{v}}^{\text{fwd}}$ and $\hat{\mathbf{v}}^{\text{bwd}}$, and supervise them using the corresponding scene flow fields:
\begin{equation}
    \mathcal{L}_{\text{flow}} = \|\hat{\mathbf{v}}^{\text{fwd}} - \mathbf{v}^{\text{fwd}}\|_1 + \|\hat{\mathbf{v}}^{\text{bwd}} - \mathbf{v}^{\text{bwd}}\|_1.
\end{equation}

\subsection{Losses}
We jointly optimize the Gaussians and motion bases using photometric, geometric, correspondence, and regularization losses. All Gaussians are fused to render image $\hat{\mathbf{I}}$, dynamic mask $\hat{\mathbf{M}}$, depth $\hat{\mathbf{D}}$, normal $\hat{\mathbf{N}}$, 3D correspondence $\hat{\bm{\mu}}_{t\to t'}$, and velocities $\hat{\mathbf{v}}^{\text{fwd}}$ and $\hat{\mathbf{v}}^{\text{bwd}}$.
The photometric loss follows previous work~\cite{kerbl3Dgaussians} and includes an additional mask loss to constrain the occupancy of dynamic Gaussians:
\begin{equation}
\begin{split}
\mathcal{L}_{\text{photo}} 
&= (1 - \lambda_{\text{ssim}})\, \| \hat{\mathbf{I}} - \mathbf{I} \|_1 
+ \lambda_{\text{ssim}}\, \text{SSIM}(\hat{\mathbf{I}}, \mathbf{I}) \\
&\quad + \lambda_{\alpha}\, \mathcal{L}_{\text{mask}}.
\end{split}
\label{eq:loss_photo}
\end{equation}
Multi-view stereo is less effective in monocular settings for accurate geometry recovery, so additional geometric priors are needed.
Following the common practice~\cite{Lei_mosca_2025_CVPR,som2024}, we use depth loss and normal consistency loss to supervise the mean and rotation of Gaussians. 
The depth loss is computed in a robust, scale- and translation-invariant manner, while the normal consistency loss employs cosine similarity:
\begin{equation}
    \mathcal{L}_{\text{geom}} = \lambda_{\text{depth}}\, \|\hat{\mathbf{D}}^{*} - \mathbf{D}^{*}\|_1 + \lambda_{\text{normal}}\, \|\mathbf{1} - \hat{\mathbf{N}}^\text{T} \cdot \mathbf{N}\|_2^2.
    \label{eq:loss_geom}
\end{equation}
Photometric and geometric loss are not enough to guide the motion of dynamic Gaussians because $\nabla\bm{\mu_i}$ is invalid when there is not enough overlap area between $g_i$ and its potential ground truth $g_i^*$ ~\cite{Xing2022drot, Xing2023ExtendedPS}. 
To address this, we introduce correspondence supervision that combines tracking and flow losses to explicitly guide the motion of dynamic Gaussians, following~\cite{Lei_mosca_2025_CVPR,som2024}.
We render 3D trajectories $\hat{\bm{\mu}}_{t\to t'}$ and supervise them with precomputed 3D tracking $\bm{\mu}_{t\to t'}$. 
\begin{equation}
    \begin{split}
    \mathcal{L}_{\text{cor}}
    &= \lambda_{\text{track}}\, \|\hat{\bm{\mu}}_{t \to t'} - \bm{\mu}_{t \to t'}\|_1 + \lambda_{\text{flow}} \mathcal{L}_{\text{flow}}.
    \end{split}
    \label{eq:loss_cor}
\end{equation}
We also apply two regularization losses. 
Poor initialization often causes Gaussians to produce needle-like artifacts~\cite{zhang2024pixelgsdensitycontrolpixelaware}, while monocular videos provide limited structural cues.
To alleviate these issues, we use scale regularization following~\cite{stearns2024marbles} to encourage isotropic Gaussians and apply regularization on $\beta^r$ to prevent rigid Gaussians from overfitting:
\begin{equation}
    \mathcal{L}_{\text{reg}} = \frac{1}{N}\sum_{i=1}^{N}{ \lambda_{\beta} \frac{1}{\beta^r_i} + \lambda_{\mathbf{s}} \text{var}(\mathbf{s}_{i})}.
    \label{eq::reg_loss}
\end{equation}
 
\begin{figure*}[t]
  \centering
  \begin{subfigure}[t]{\textwidth}
    \centering
    \includegraphics[width=0.16\textwidth]{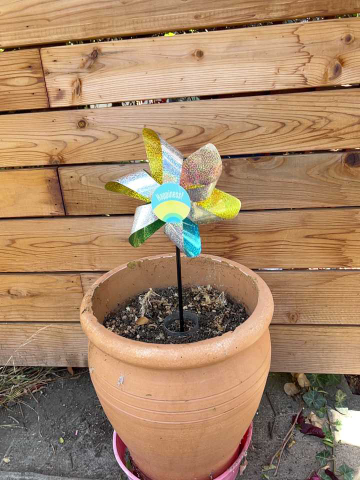}
    \includegraphics[width=0.16\textwidth]{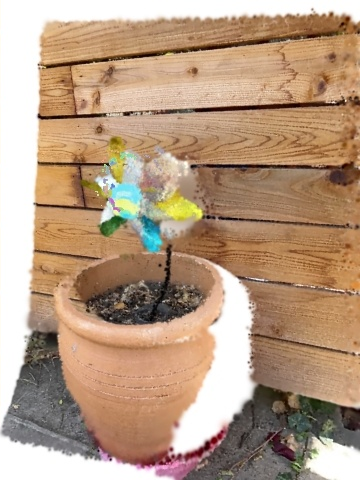}
    \includegraphics[width=0.16\textwidth]{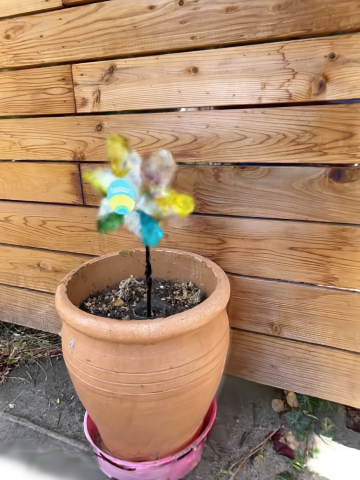}
    \includegraphics[width=0.16\textwidth]{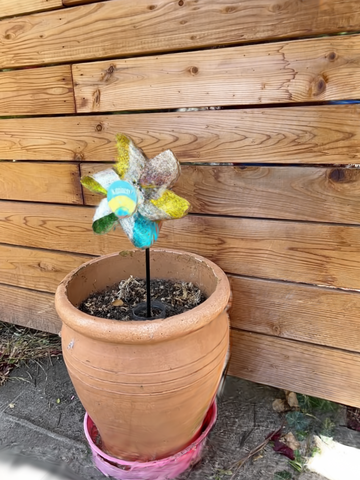}
    \includegraphics[width=0.16\textwidth]{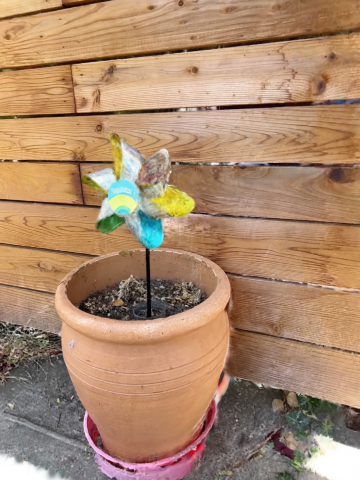}
    \includegraphics[width=0.16\textwidth]{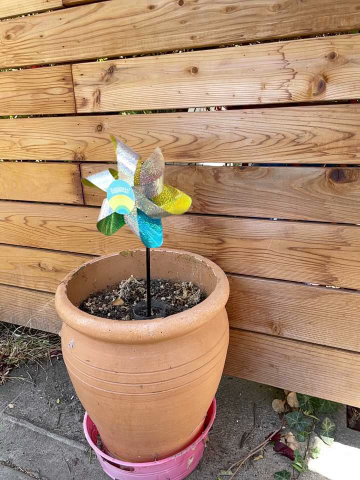}
  \end{subfigure}
  
  \vspace{4pt}
  
  \begin{subfigure}[t]{\textwidth}
    \centering
    
    \begin{subfigure}[t]{0.16\textwidth}
      \centering
      \includegraphics[width=\linewidth]{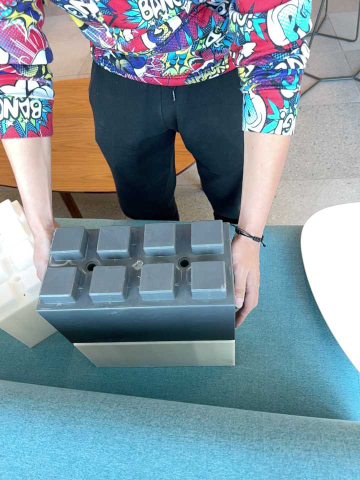}
      \caption*{\footnotesize Input View}
    \end{subfigure}
    \begin{subfigure}[t]{0.16\textwidth}
      \centering
      \includegraphics[width=\linewidth]{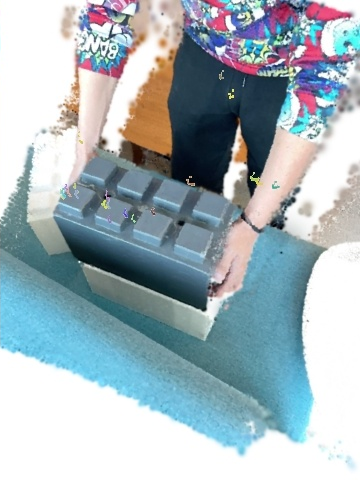}
      \caption*{\footnotesize Gaussian Marbles~\cite{stearns2024marbles}}
    \end{subfigure}
    \begin{subfigure}[t]{0.16\textwidth}
      \centering
      \includegraphics[width=\linewidth]{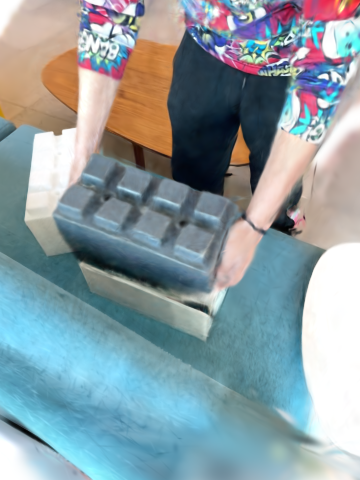}
      \caption*{\footnotesize MoSca~\cite{Lei_mosca_2025_CVPR}}
    \end{subfigure}
    \begin{subfigure}[t]{0.16\textwidth}
      \centering
      \includegraphics[width=\linewidth]{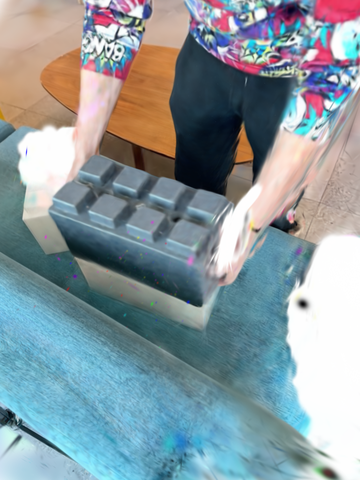}
      \caption*{\footnotesize Shape of Motion~\cite{som2024}}
    \end{subfigure}
    \begin{subfigure}[t]{0.16\textwidth}
      \centering
      \includegraphics[width=\linewidth]{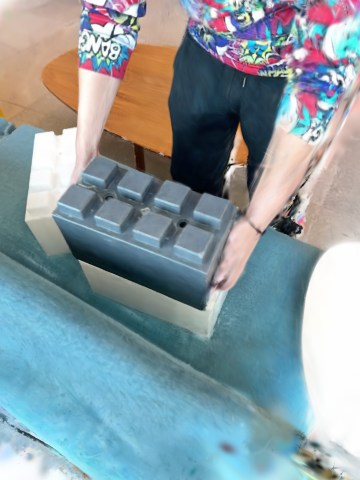}
      \caption*{\footnotesize Ours}
    \end{subfigure}
    \begin{subfigure}[t]{0.16\textwidth}
      \centering
      \includegraphics[width=\linewidth]{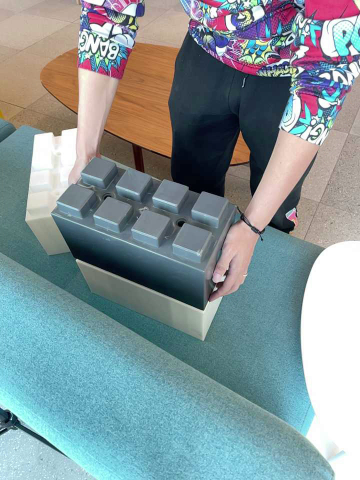}
      \caption*{\footnotesize Ground Truth}
    \end{subfigure}
  \end{subfigure}
  
  \caption{
  \textbf{Evaluation on DyCheck iPhone Dataset.}
Our method is comparable to the previous SOTA method~\cite{Lei_mosca_2025_CVPR} on the DyCheck dataset.
We present qualitative comparisons on the \textit{paper-windmill} and \textit{block} scenes.
Gaussian Marbles~\cite{stearns2024marbles} tends to overfit to input views.
MoSca~\cite{Lei_mosca_2025_CVPR} models the motion in the paper-windmill scene effectively, yet produces blurry surfaces on the object.
In addition, MoSca computes dynamic masks using the Sampson error, which leads to incorrect segmentation in the \textit{block} scene.
Compared with SoM~\cite{som2024}, our method achieves similar performance in modeling dynamic objects while reconstructing static regions more robustly.
  }
  \label{fig:qualitative_dycheck}
  \vspace{-10pt}
\end{figure*}
\section{Experiments}
We evaluate the proposed method on the \textbf{Nvidia Dynamic Scenes Dataset}~\cite{Yoon2020NovelVS} and \textbf{Dycheck iPhone Dataset}~\cite{gao2022dynamic}. 
We report quantitative results of novel view synthesis using PSNR, SSIM, and LPIPS~\cite{zhang2018perceptual} as evaluation metrics.
To further demonstrate the effectiveness of our representation, we also experiment on selected scenes from the DAVIS dataset~\cite{DAVIS2016, DAVIS2017}, which contain fast motion and large non-rigid deformations.
Finally, we conduct ablation studies on the Nvidia Dynamic Scenes dataset to analyze the contribution of each proposed component.

\subsection{Implementation Details}
All experiments are conducted on a single NVIDIA A100 GPU.
For in-the-wild videos, we obtain camera parameters and metric depth following~\cite{huang2025vipe}, with the depth model set to MoGe~\cite{wang2025moge}.
For evaluation, we follow the processing pipeline of the current state-of-the-art method~\cite{Lei_mosca_2025_CVPR}. 
Specifically, for the Nvidia Dynamic Scenes dataset, we use Metric3D~\cite{yin2023metric3d} to obtain metric depth, while for the Dycheck iPhone dataset, we use its provided LiDAR depth.
For both datasets, we use the MoCa module~\cite{Lei_mosca_2025_CVPR} to obtain depth-aligned camera parameters. We use SEA-RAFT~\cite{raft} for optical flow estimation, BootsTAP~\cite{doersch2024bootstap} for 2D point tracking, and SAM2~\cite{ravi2024sam2segmentimages} for object segmentation.

During training, all Gaussians and motion bases are optimized by the Adam optimizer~\cite{kingma2017adammethodstochasticoptimization}. Camera poses are not optimized to speed up the training. 
The total number of training iterations is set to 30K for the Nvidia Dynamic
Scenes dataset and 100K for the Dycheck iPhone dataset, each with a batch size of 1.
We adopt gsplat~\cite{ye2025gsplat} for fast multi-feature rendering. We use 2DGS~\cite{Huang2DGS2024} as our rendering backend to better leverage the depth prior. The number of motion bases $K$ is set to 10 for all scenes to ensure a fair comparison.
The temporal parameters $t$, $\beta$, and $\gamma$ are all defined in frame index units. Since our method is a per-scene optimization-based approach, the choice of time unit does not strongly affect the final results.
The sharpness $\alpha$ of the soft gating function is set to 3.0, and $\beta_r$ is set to 2.0.
Training a 270-frame video at a resolution of $480 \times 360$(30K iterations) takes approximately 30 minutes, and the rendering speed is around 160 FPS.

\subsection{Evaluation on Novel View Synthesis}

\noindent\textbf{Nvidia Dynamic Scenes Dataset.} We follow the protocol in DynamicNeRF~\cite{gao2021dynamicviewsynthesisdynamic} to conduct novel view synthesis evaluation experiments on the Nvidia dataset. We report PSNR and LPIPS in Table~\ref{tab:Nvidia}. We further show the qualitative comparison in Figure~\ref{fig:qualitative_nvidia}.

As shown in Table~\ref{tab:Nvidia}, our method outperforms both the 3DGS-based SOTA~\cite{Lei_mosca_2025_CVPR} and NeRF-based SOTA~\cite{Gao-ICCV-DynNeRF}, which illustrates the superiority of our method. 
As shown in Figure~\ref{fig:qualitative_nvidia}, our method produces significantly fewer artifacts than MoSca~\cite{Lei_mosca_2025_CVPR} in the \textit{Umbrella} scene, where MoSca relies on node-interpolated $\mathbb{SE}(3)$ transformations to drive Gaussian motion. Our representation allows rigid and transient Gaussians to appear and disappear adaptively, maintaining a clean and compact distribution around deforming surfaces and avoiding the accumulation of unobserved Gaussians.
In the \textit{Playground} scene, MoSca fails to reconstruct the woman’s facial expression in novel views, while our method preserves fine-grained details. We attribute this improvement to two factors. First, our object-wise dynamic mask is more complete and temporally consistent, preventing static Gaussians from contaminating dynamic regions. Second, facial expressions involve subtle, highly non-rigid, and high-frequency motions that $\mathbb{SE}(3)$-driven models inherently struggle to capture. RiGS models these fine-scale deformations through transient Gaussians, enabling accurate reconstruction of such delicate dynamics.

\noindent\textbf{Dycheck iPhone Dataset.} The Dycheck iPhone dataset consists of 14 causally captured scenes with lengths ranging from 200 to 500 frames. Each scene has one monocular video with a handheld moving camera as reconstruction input and up to two extreme static captures as evaluation ground truth.  
The motions in these scenes are generally smooth and sometimes regular, making them relatively suitable for reconstruction under the as-rigid-as-possible assumption~\cite{Lei_mosca_2025_CVPR, som2024}.
We report mPSNR, mSSIM, and mLPIPS following the protocol used in ~\cite{Lei_mosca_2025_CVPR, gao2022dynamic} in Table~\ref{tab:dycheck}.  
Before evaluation, we perform test-time pose alignment to compensate for inaccuracies in the provided camera poses.
Our method achieves comparable results and outperforms it in the mPSNR and mLPIPS metrics, demonstrating the superior perceptual quality of our reconstruction.
We further show the quantitative comparison on the Dycheck iPhone Dataset in Figure~\ref{fig:qualitative_dycheck}.
\begin{table}
  \caption{Quantitative Evaluation on Dycheck Dataset}
  \label{tab:dycheck}
  \centering
  \setlength{\tabcolsep}{2pt}
  \begin{tabular}{@{}lcccccccc@{}}
    \toprule
    Methods & mPSNR$\uparrow$ & mSSIM$\uparrow$ & mLPIPS$\downarrow$\\
    \midrule
    T-NeRF~\cite{gao2022dynamic} & 16.96 & 0.577 & 0.379 \\
    HyperNeRF~\cite{park2021hypernerf} & 16.81 & 0.569 & 0.332 \\
    DynPoint~\cite{zhou2025dynpointdynamicneuralpoint} & 16.89 & 0.573 & - \\
    Gaussian Marbles~\cite{stearns2024marbles} & 16.72 & - & 0.418 \\
    Shape-of-Motion~\cite{som2024} & \third{17.32} & \third{0.598} & \third{0.296} \\
    MoSca~\cite{Lei_mosca_2025_CVPR} & \second{19.32} & \first{0.706} & \second{0.264} \\
    Ours & \first{19.50} & \second{0.705} & \first{0.240} \\
    \bottomrule
  \end{tabular}
\end{table}
\subsection{Comparison on Extreme Motion Scenes}
\begin{figure}[t]
  \centering
  \begin{subfigure}[t]{\columnwidth}
    \centering
    \includegraphics[width=0.24\columnwidth]{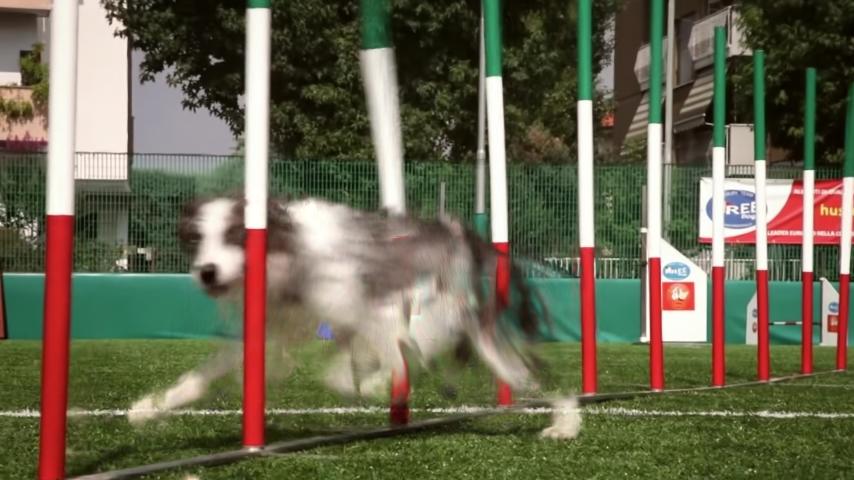}
    \includegraphics[width=0.24\columnwidth]{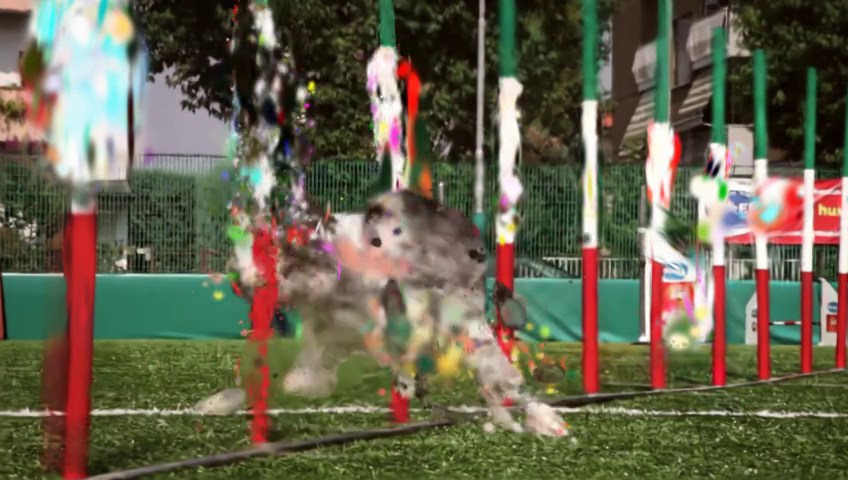}
    \includegraphics[width=0.24\columnwidth]{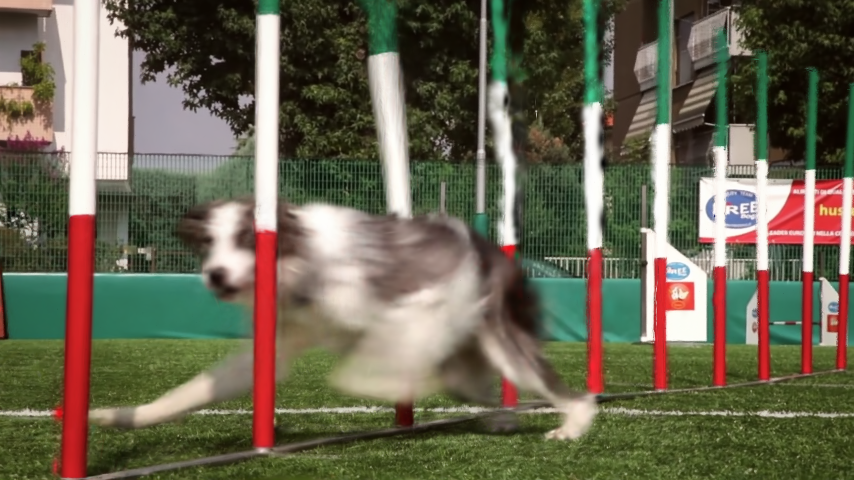}
    \includegraphics[width=0.24\columnwidth]{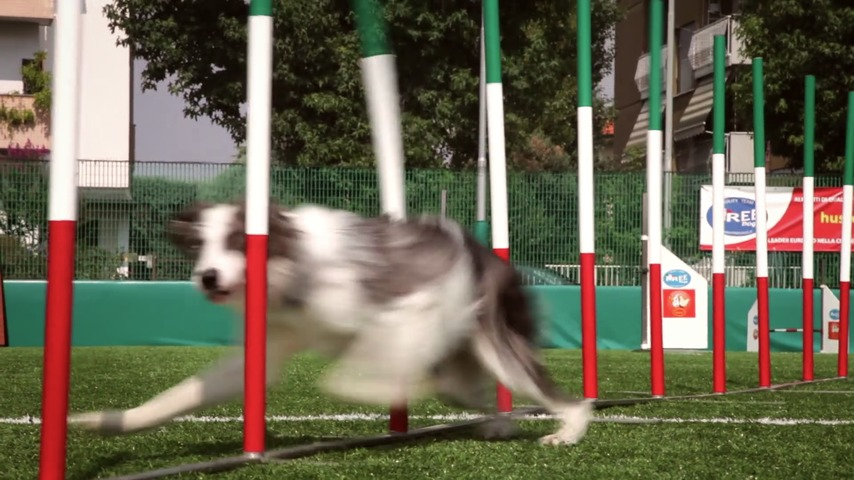}
  \end{subfigure}

  \vspace{4pt}

  \begin{subfigure}[t]{\columnwidth}
    \centering

    \begin{subfigure}[t]{0.24\columnwidth}
      \centering
      \includegraphics[width=\linewidth]{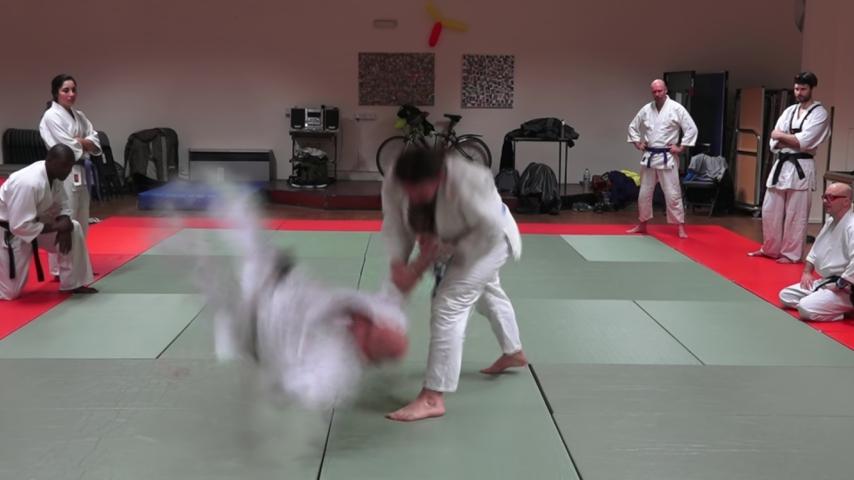}
      \caption*{\footnotesize MoSca~\cite{Lei_mosca_2025_CVPR}}
    \end{subfigure}
    \begin{subfigure}[t]{0.24\columnwidth}
      \centering
      \includegraphics[width=\linewidth]{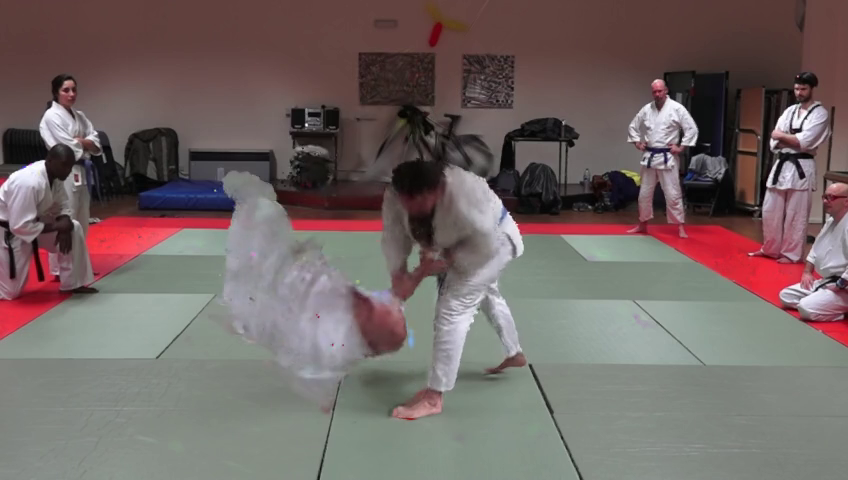}
      \caption*{\footnotesize SoM~\cite{som2024}}
    \end{subfigure}
    \begin{subfigure}[t]{0.24\columnwidth}
      \centering
      \includegraphics[width=\linewidth]{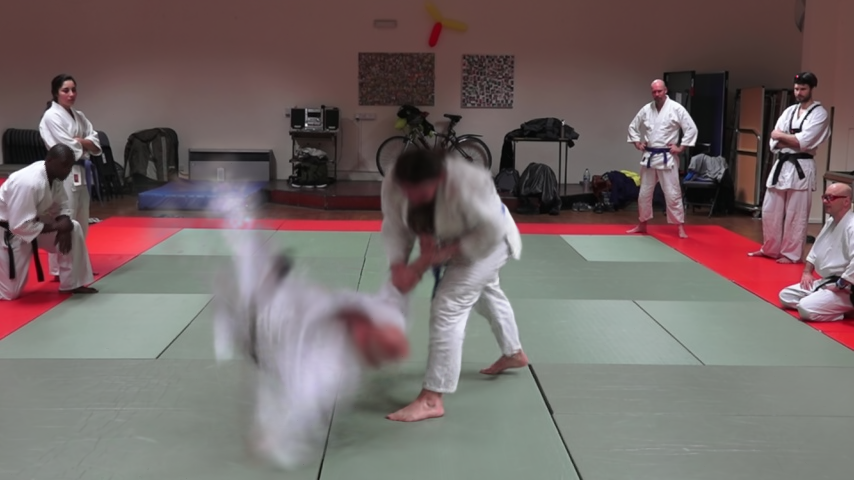}
      \caption*{\footnotesize Ours}
    \end{subfigure}
    \begin{subfigure}[t]{0.24\columnwidth}
      \centering
      \includegraphics[width=\linewidth]{images/judo/judo_gt.png}
      \caption*{\footnotesize Ground Truth}
    \end{subfigure}
  \end{subfigure}

  \caption{Qualitative Comparison on Extreme Motion Scenes.}
  \label{fig:extreme_motion}
\end{figure}

The main advantage of our method lies in its ability to model multi-scale motion dynamics within a unified framework. While the motion patterns in the DyCheck iPhone dataset are mostly smooth—and scenes such as paper-windmill and spin exhibit predominantly regular motions—they do not fully reveal the benefits of multi-scale motion modeling. To more comprehensively evaluate this capability, we further compare our approach with Shape of Motion~\cite{som2024} and MoSca~\cite{Lei_mosca_2025_CVPR} on the judo and dog-agility scenes from the DAVIS dataset~\cite{DAVIS2016, DAVIS2017}, which contain rapid and complex object movements.
As shown in Figure~\ref{fig:extreme_motion}, both MoSca and Shape of Motion fail to recover the fast-moving regions, whereas our method successfully captures these challenging dynamics.

\begin{table}[t]
\centering
\scriptsize
\setlength{\tabcolsep}{3pt}
\renewcommand{\arraystretch}{1.15}

\caption{Ablation Results on the Nvidia Dynamic Scene Dataset.}
\begin{tabular}{lccccccc}
\toprule
\textbf{Methods} & Mask & Scene Flow & Trans. & Rigid & PSNR$\uparrow$ & SSIM$\uparrow$ & LPIPS$\downarrow$\\
\midrule
Ours (Full)& \checkmark & \checkmark & \checkmark & \checkmark & \textbf{27.43} & \textbf{0.879} & \textbf{0.051} \\
Sampson Mask & & \checkmark & \checkmark & \checkmark & 26.00 & 0.869 & 0.069\\
wo. Scene Flow & \checkmark &   & \checkmark & \checkmark & 27.27& 0.877 &0.052 \\
wo. Rigid Gaus. & \checkmark & \checkmark& \checkmark &  & 27.11&0.876 &0.055 \\
wo. Trans. Gaus. & \checkmark & \checkmark &  & \checkmark & 26.87&0.867 & 0.063 \\
\bottomrule
\end{tabular}
\label{tab:ablation}
\end{table}

\subsection{Ablation Study}
We conduct ablation studies on the key components of our framework, and the results are reported in Table~\ref{tab:ablation}.
We first validate the effectiveness of the proposed object-wise dynamic masks by replacing them with Sampson-based masks following~\cite{liu2023robust}. This substitution leads to a drastic performance drop. The reason is that rigid Gaussians are initialized from tracks within the dynamic masks; when the masks are inaccurate, rigid-Gaussian initialization fails, causing the optimization to rely almost entirely on static Gaussians and resulting in severe degradation.
We then remove the scene flow loss to examine its contribution. Without this constraint, the predicted scene flow becomes noticeably noisier and the rendering quality deteriorates accordingly.
Finally, to evaluate the benefit of our mixed representation, we train rigid Gaussians and transient Gaussians in isolation. Removing rigid Gaussians causes large empty regions, while removing transient Gaussians significantly reduces fine details, 
showing the complementary roles of the two components and the importance of their joint optimization.
\section{Conclusion}

In this work, we have presented RiGS, a rigid-aware 4D Gaussian Splatting framework that reconstructs dynamic 3D scenes from monocular videos by explicitly modeling motion across multiple temporal scales. By decomposing the scene into static, rigid, and transient Gaussian primitives, RiGS effectively captures long-term smooth motions as well as short-term high-frequency dynamics. The proposed object-wise dynamic mask enables reliable separation of static and dynamic regions, while the transition mechanism between rigid and transient Gaussians, together with scene-flow–guided optimization, provides dense and coherent 3D motion supervision. 
Extensive experiments on diverse benchmarks demonstrate that RiGS achieves state-of-the-art performance in reconstruction quality and temporal consistency, paving the way toward more consistent and temporally coherent dynamic scene modeling.

\section*{Acknowledgements} 
This work was supported in part by the NIH grant R01HD104969 and the NTU Nanyang Assistant Professorship Startup Grant 025661-00012. This research was also supported in part by Google.org and the Google Cloud Research Credits Program through the Gemma Academic Program.

{
    \small
    \bibliographystyle{ieeenat_fullname}
    \bibliography{main}
}

\clearpage
\setcounter{page}{1}
\maketitlesupplementary
\begin{table}[t]
  \caption{Hyper Parameters}
  \label{tab:hyper_param_only}
  \centering
  \setlength{\tabcolsep}{6pt}
  \begin{tabular}{@{}lc|lc@{}}
    \toprule
    \textbf{Parameter} & \textbf{Value} &
    \textbf{Parameter} & \textbf{Value} \\
    \midrule
    $\lambda_{\text{ssim}}$ & 0.1 &
    $\text{lr}_{\bm\mu}$ & 0.00016 \\

    $\lambda_{\text{alpha}}$ & 0.5 &
    $\text{lr}_{\mathbf{s}}$ & 0.005 \\

    $\lambda_{\text{depth}}$ & 0.05 &
    $\text{lr}_{\mathbf{q}}$ & 0.001 \\

    $\lambda_{\text{normal}}$ & 0.05 &
    $\text{lr}_{o}$ & 0.05 \\

    $\lambda_{\text{track}}$ & 2.0 &
    $\text{lr}_{\mathbf{c}}$ & 0.01 \\

    $\lambda_{\text{flow}}$ & 0.01 &
    $\text{lr}_{\beta}$ & 0.001 \\

    $\lambda_{\beta}$ & 0.5 &
    $\text{lr}_{\gamma}$ & 0.001 \\

    $\lambda_{s}$ & 0.5 &
    $\text{lr}_{\mathbf{w}}$ & 0.01 \\

    --- & --- &
    $\text{lr}_{\mathbf{T}}$ & 0.0001 \\
    \bottomrule
  \end{tabular}
\end{table}
\section{Additional Implementation Details}
\noindent\textbf{Object-Wise Dynamic Masks.}
As shown in Eq.~\ref{eq:score}, we compute motion scores by combining the flow-based weights $w_t$ with the Sampson error~\cite{sampson}. To obtain $w_t$, we use the flow uncertainty $u_t \in \mathbb{R}^+$ estimated by SEA-RAFT~\cite{raft} together with the occlusion mask $m_t^{\text{occ}}$ from a forward–backward consistency check~\cite{meister2017unflow}:
\begin{equation}
w_t = \frac{1 - m_t^{\text{occ}}}{(1 + u_t)^2}.
\label{eq:flow_weight}
\end{equation}

The Sampson error is given by
\begin{equation}
e =
\frac{
\left| \mathbf{x}_{\text{l}}^{\mathrm{T}} \mathbf{F} \mathbf{x}_{\text{r}} \right|
}{
\sqrt{ \| \mathbf{F} \mathbf{x}_{\text{l}} \|_2^2 + \| \mathbf{F} \mathbf{x}_{\text{r}} \|_2^2 }
},
\label{eq:sampson}
\end{equation}
where $\mathbf{x}_{\text{l}}$ and $\mathbf{x}_{\text{r}}$ are sampled from the non-occluded regions $\neg m^{\text{occ}}$. The fundamental matrix $\mathbf{F}$ is estimated using the LMEDS (Least Median of Squares) method based on 10,000 sampled correspondences for efficiency.

For object mask generation, we apply SAM~\cite{ravi2024sam2segmentimages, kirillov2023segany} for automatic video object segmentation. To detect objects that do not appear in the first frame, we modify AutoSeg to enable fast discovery of new object instances. For temporal filtering, we set $\epsilon^{\text{temp}} = 1\mathrm{e}{-4}$ as a conservative threshold for static objects. Since different videos exhibit varying FPS and motion magnitudes, the dynamic threshold $\epsilon^{\text{dyn}}$ is adaptively set to $\max(s_i)/4$.

\begin{figure}
    \centering
    \includegraphics[width=1\linewidth]{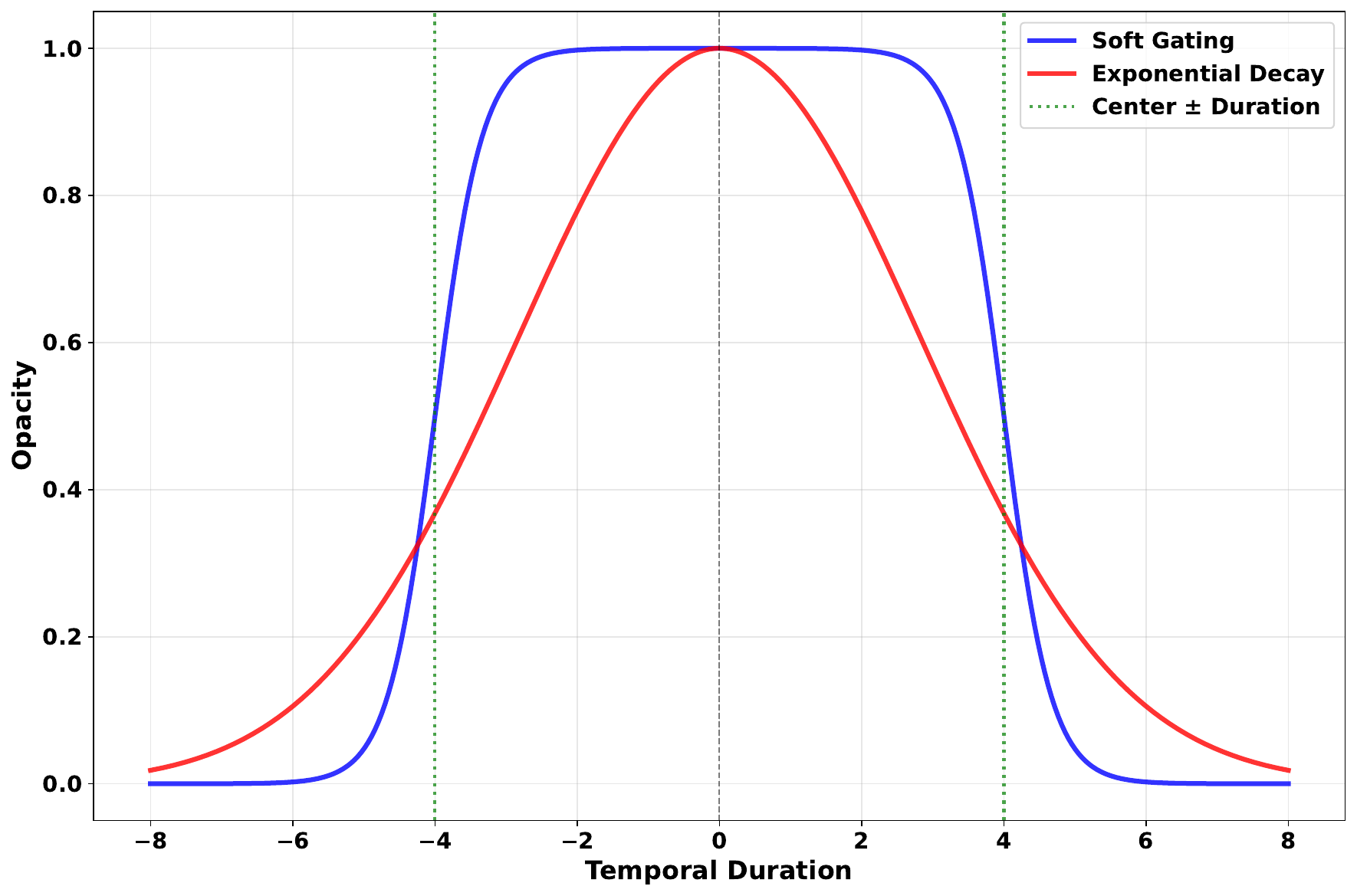}
    \caption{Comparison of Soft Gating and Exponential Decay Functions}
    \label{fig:func_compare}
\end{figure}

\noindent\textbf{Hyper-Parameters}
We present additional hyper-parameters in Table~\ref{tab:hyper_param_only}. All kinds of Gaussians share the same set of learning rates. For the Nvidia dataset, we use 3K iterations for static Gaussian warm-up and 12K iterations for rigid Gaussian warm-up, while for the DyCheck iPhone dataset, these values are set to 5K and 40K, respectively.

\noindent\textbf{Opacity Decay}
While prior works~\cite{yang2023gs4d, luo2025instant4d, wang2025freetimegs} use an exponential distribution for opacity decay, we employ a soft gating function due to its sharper temporal boundary. We compare the two functions in Figure~\ref{fig:func_compare}. As shown in the figure, the soft gating function maintains a flat response near its peak, whereas the exponential decay decreases gradually. We also experiment with replacing our soft gating function with the exponential form and observe that it leads to the emergence of more short-term Gaussians. We attribute this behavior to the soft boundary of the exponential decay when the temporal duration becomes large.

\begin{figure*}[t]
  \centering
  \begin{subfigure}[t]{\textwidth}
    \centering
    \includegraphics[width=0.24\textwidth]{images/dist/judo_colors.png}
    \includegraphics[width=0.24\textwidth]{images/dist/fish_colors.png}
    \includegraphics[width=0.24\textwidth]{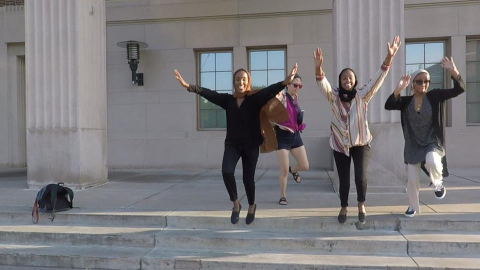}
    \includegraphics[width=0.24\textwidth]{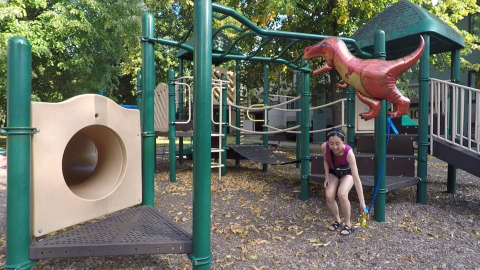}
  \end{subfigure}

  \vspace{4pt}
  \begin{subfigure}[t]{\textwidth}
    \centering
    \includegraphics[width=0.24\textwidth]{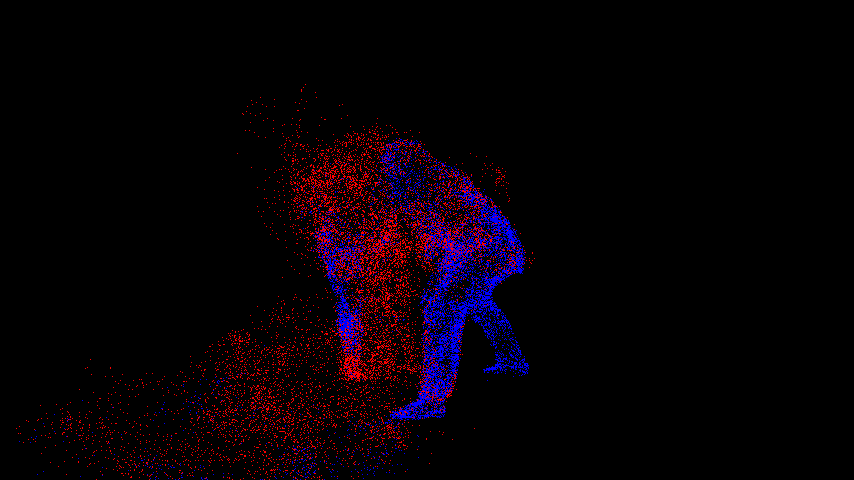}
    \includegraphics[width=0.24\textwidth]{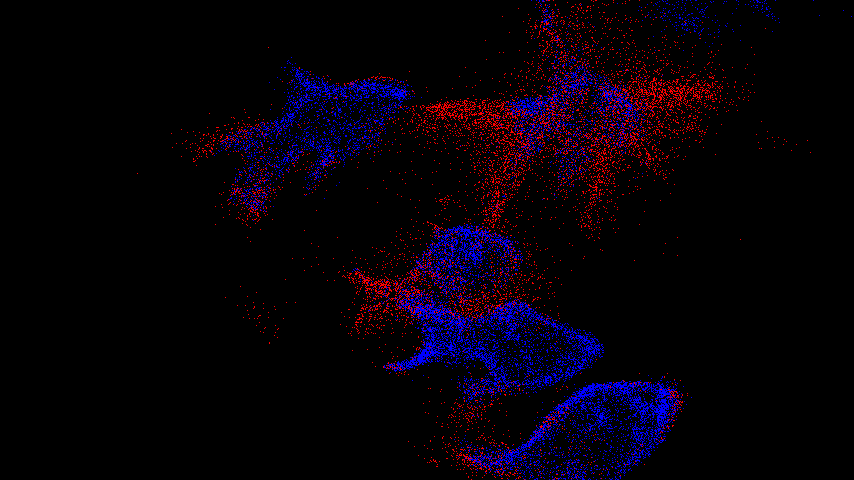}
    \includegraphics[width=0.24\textwidth]{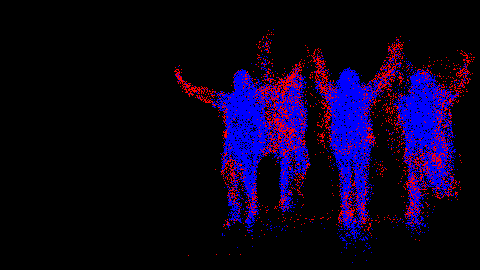}
    \includegraphics[width=0.24\textwidth]{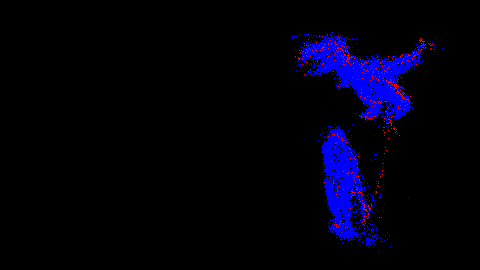}
  \end{subfigure}
  \vspace{4pt}

  \begin{subfigure}[t]{\textwidth}
    \centering
    
    \begin{subfigure}[t]{0.24\textwidth}
      \centering
      \includegraphics[width=\linewidth]{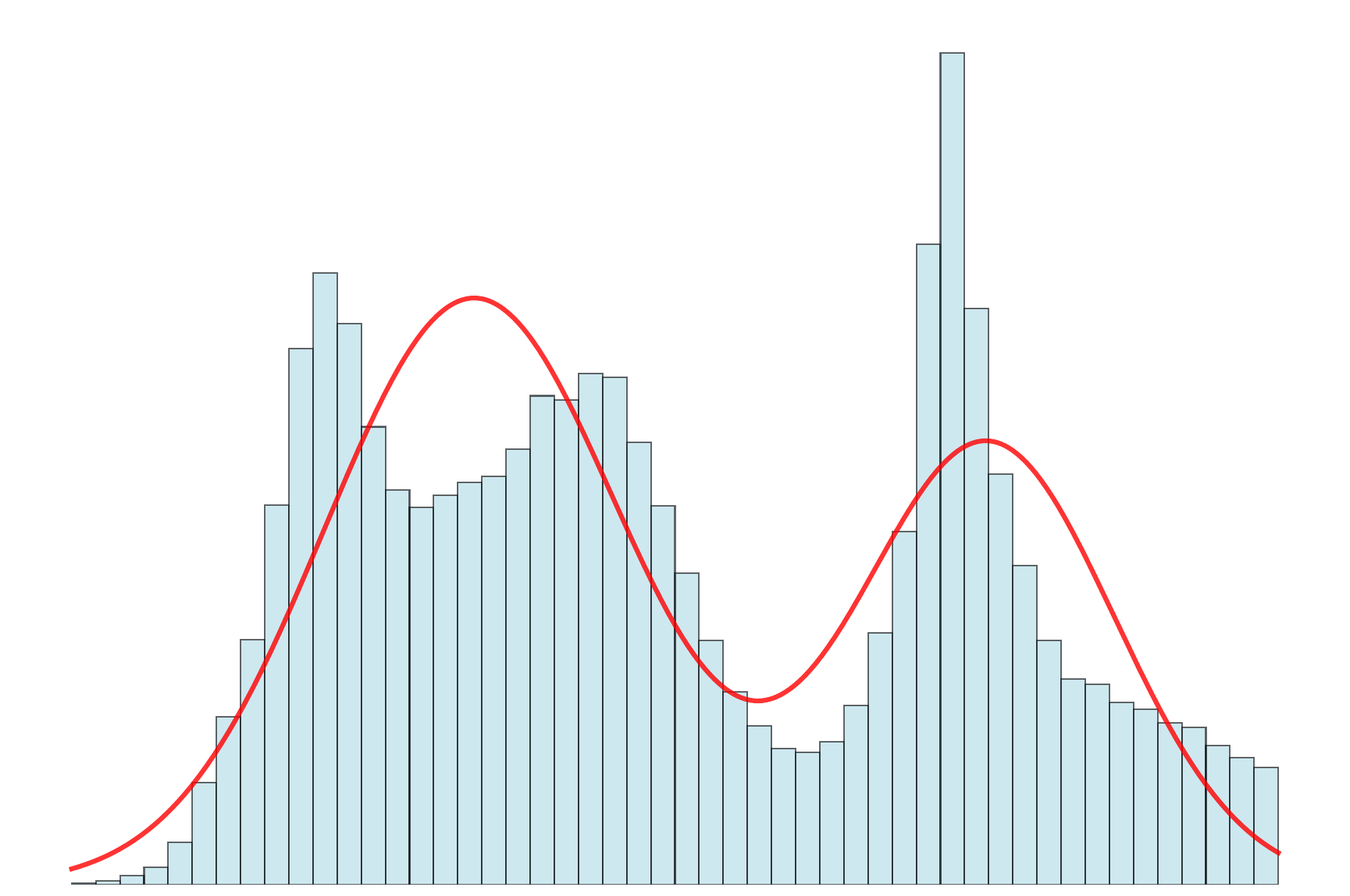}
      \caption*{\footnotesize Judo}
    \end{subfigure}
    \begin{subfigure}[t]{0.24\textwidth}
      \centering
      \includegraphics[width=\linewidth]{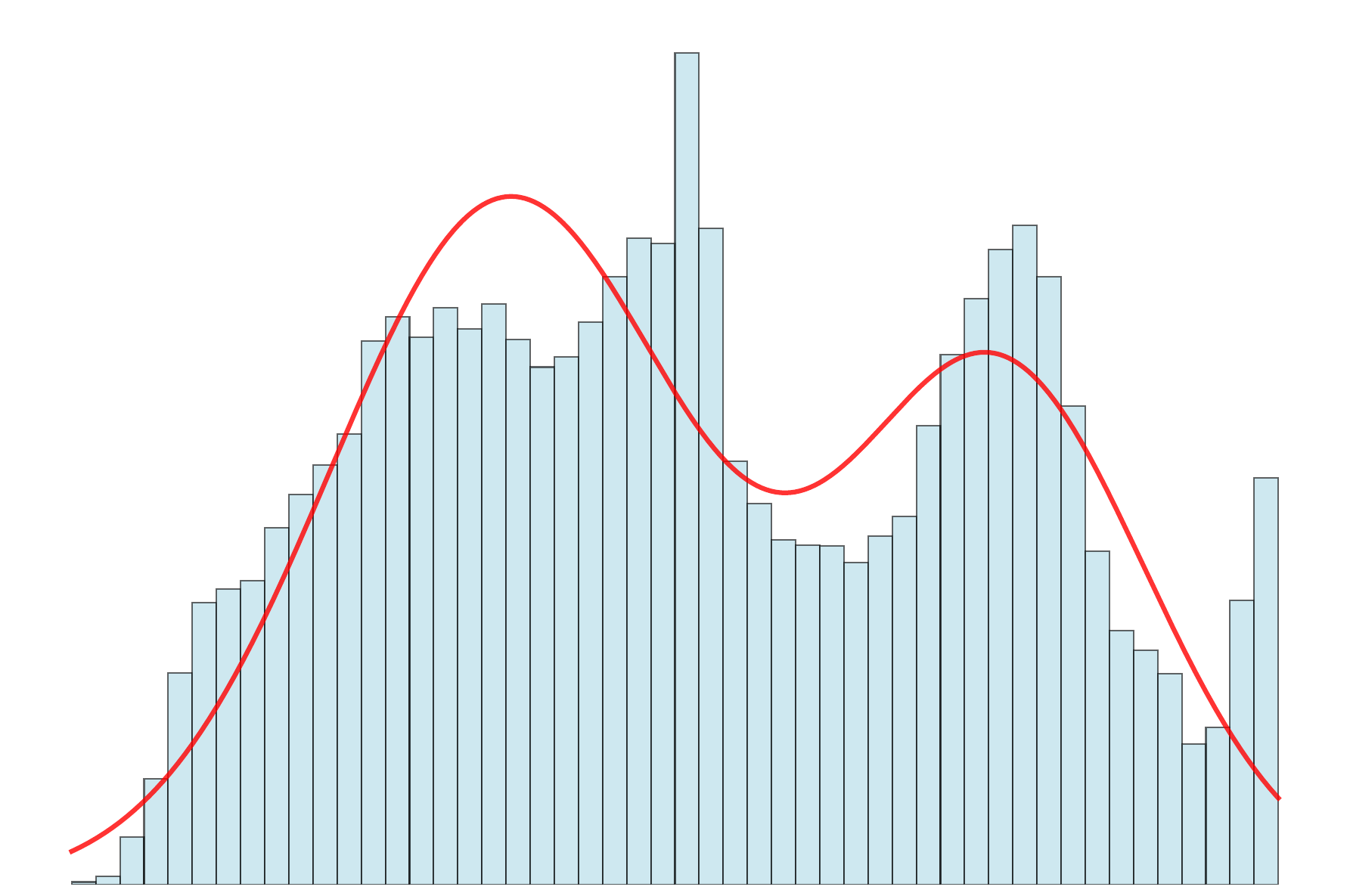}
      \caption*{\footnotesize Gold-fish}
    \end{subfigure}
    \begin{subfigure}[t]{0.24\textwidth}
      \centering
      \includegraphics[width=\linewidth]{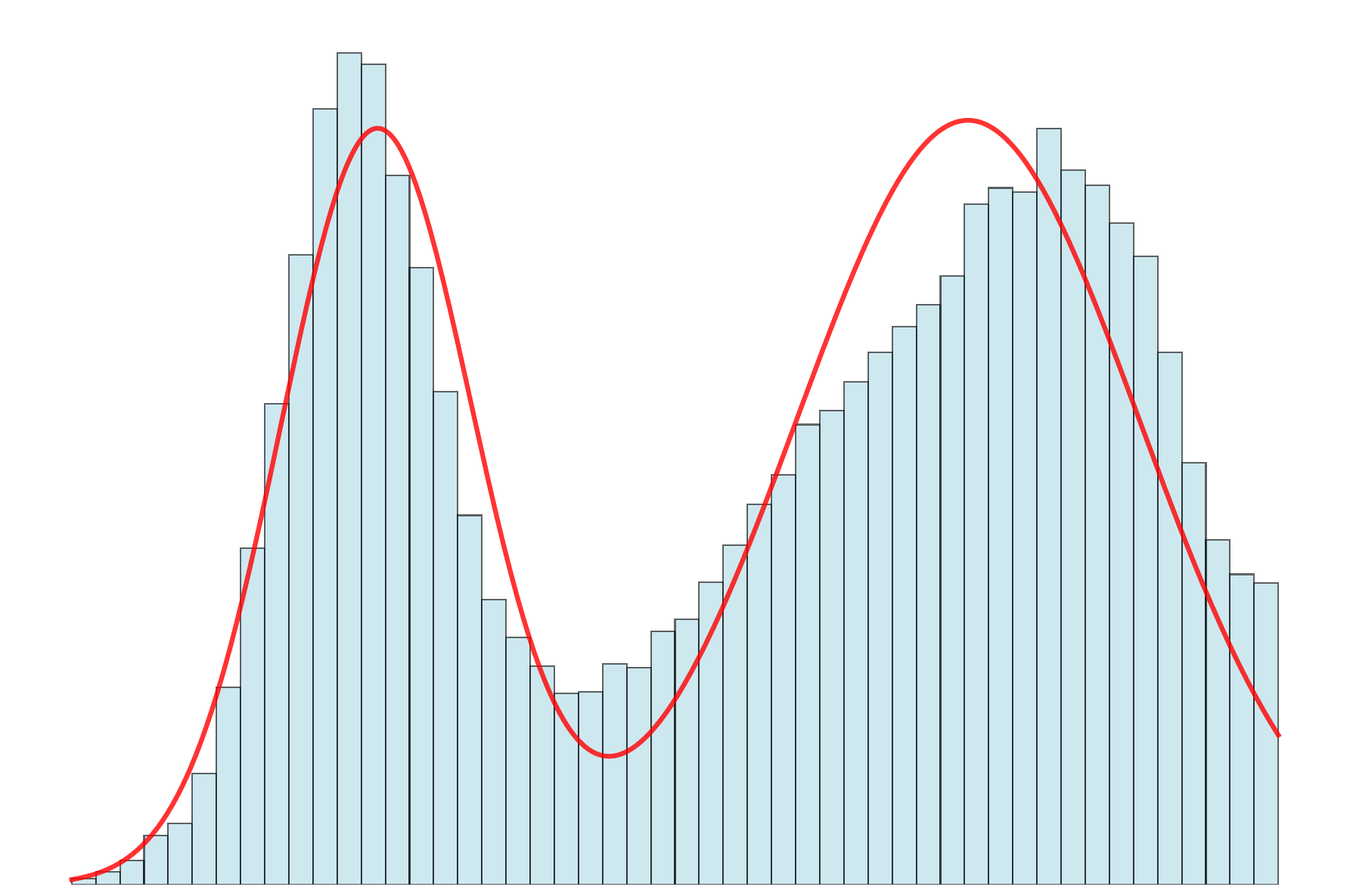}
      \caption*{\footnotesize Jumping}
    \end{subfigure}
    \begin{subfigure}[t]{0.24\textwidth}
      \centering
      \includegraphics[width=\linewidth]{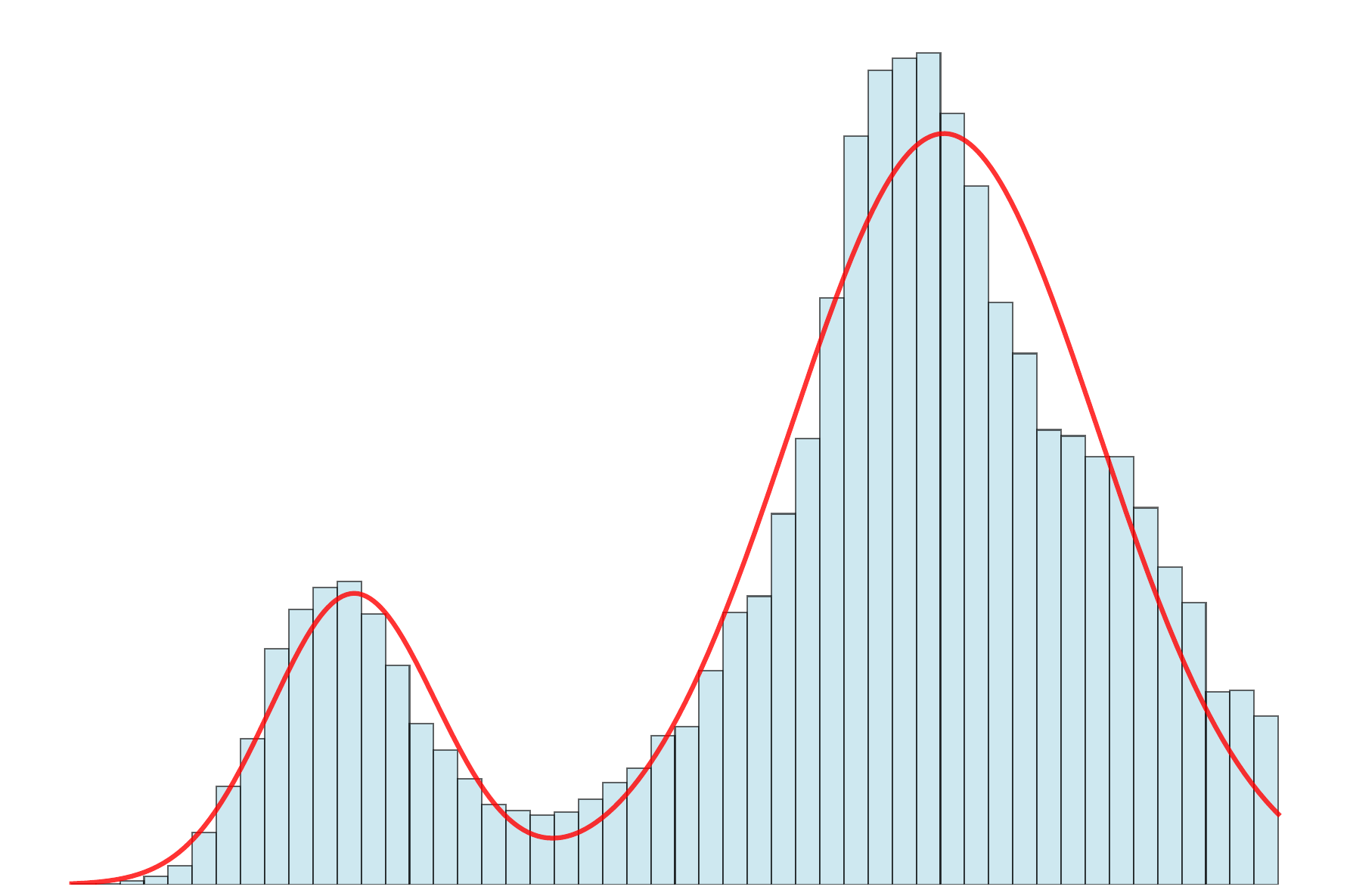}
      \caption*{\footnotesize Playground}
    \end{subfigure}
    
  \end{subfigure}
  \caption{
    \textbf{Distribution of Temporal Duration.} We visualize the distribution of the temporal duration $\beta^r$. The first row shows the original image of the scene. The second row visualizes the canonical points of the rigid Gaussians (blue) and transient Gaussians (red). The third row presents the corresponding statistical distribution.
  }
  \label{fig:temporal_vis}
  \vspace{-10pt}
\end{figure*}

\section{Two-peak Pattern}
To verify that the observed two-peak pattern is not tied to a particular scene, we further sample sequences from both the Nvidia dataset~\cite{gao2021dynamicviewsynthesisdynamic} and DAVIS~\cite{DAVIS2016, DAVIS2017}. As shown in Figure~\ref{fig:temporal_vis}, transient Gaussians predominantly correspond to fast or complex motions, whereas rigid Gaussians align with more stable, consistent motions.

We attribute this separation to the regularization term in Eq.~\ref{eq::reg_loss}. When certain Gaussians are occluded in the training frames, the regularization encourages them to remain in the scene and follow the motion trajectories represented by the motion bases. In such cases, their temporal duration $\beta^r$ tends to expand, as these Gaussians continue to exhibit coherent motion over time.

Conversely, in regions with fast or highly nonlinear motion, the limited expressiveness of the motion bases causes rigid Gaussians to drift into unwanted areas in other frames, making them appear as artifacts. The temporal duration penalty suppresses these unstable Gaussians by reducing their $\beta^r$, effectively turning them into transient Gaussians. Meanwhile, for rigid Gaussians that correctly model stable or occluded regions, the regularization reinforces long-term consistency and preserves large temporal durations.

Together, these opposing behaviors naturally induce the two-peak temporal-duration distribution, with one peak corresponding to transient Gaussians that model rapid or complex motion, and the other to rigid Gaussians that capture stable and persistent motion patterns.

\begin{table}
  \caption{Runtime Profile}
  \label{tab:time_eval}
  \centering
  \begin{tabular}{@{}lc@{}}
    \toprule
    Step & Time(s)$\downarrow$\\
    \midrule
    Optical Flow Estimation & 1.763\\
    Video Object Segmentation & 23.151 \\
    Sampson Error Calculation & 8.308 \\
    Mask Composition & 0.911 \\
    \midrule
    All & 34.133 \\
    \bottomrule
  \end{tabular}
\end{table}

\begin{table}
  \caption{Evaluation on DAVIS 2016 Dataset}
  \label{tab:davis_eval}
  \centering
  \begin{tabular}{@{}lcc@{}}
    \toprule
    Methods & IoU$\uparrow$ & Time(s)$\downarrow$\\
    \midrule
    RoMo~\cite{golisabour2024romo} & 65.9 & 552.7 \\
    SegAnyMo~\cite{Huang_2025_CVPR_seganymo} & \textbf{85.3} & 176.3 \\
    Ours & 81.9 & \textbf{34.1} \\
    \bottomrule
  \end{tabular}
\end{table}
\begin{figure}[t]
  \begin{subfigure}[t]{\columnwidth}
    \centering

    \begin{subfigure}[t]{0.24\columnwidth}
      \centering
      \includegraphics[width=\linewidth]{images/surf/image.jpg}
      \caption*{\footnotesize Kite-surf}
    \end{subfigure}
    \begin{subfigure}[t]{0.24\columnwidth}
      \centering
      \includegraphics[width=\linewidth]{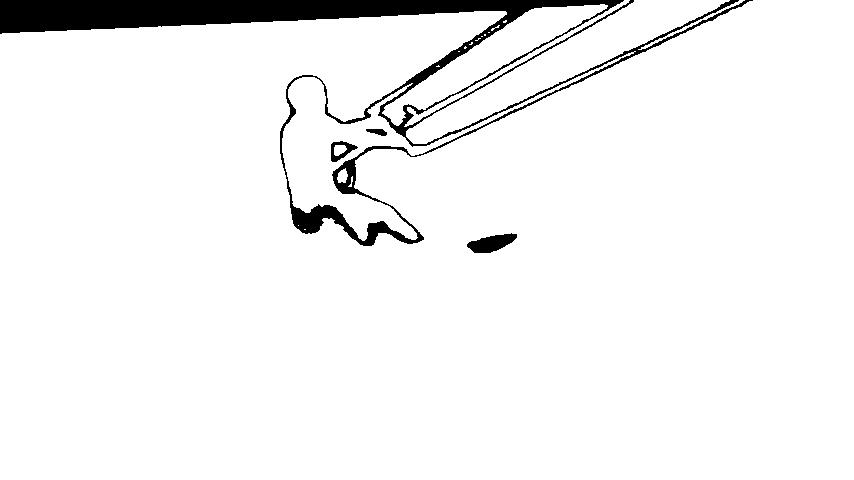}
      \caption*{\footnotesize SegAnyMo~\cite{kirillov2023segany}}
    \end{subfigure}
    \begin{subfigure}[t]{0.24\columnwidth}
      \centering
      \includegraphics[width=\linewidth]{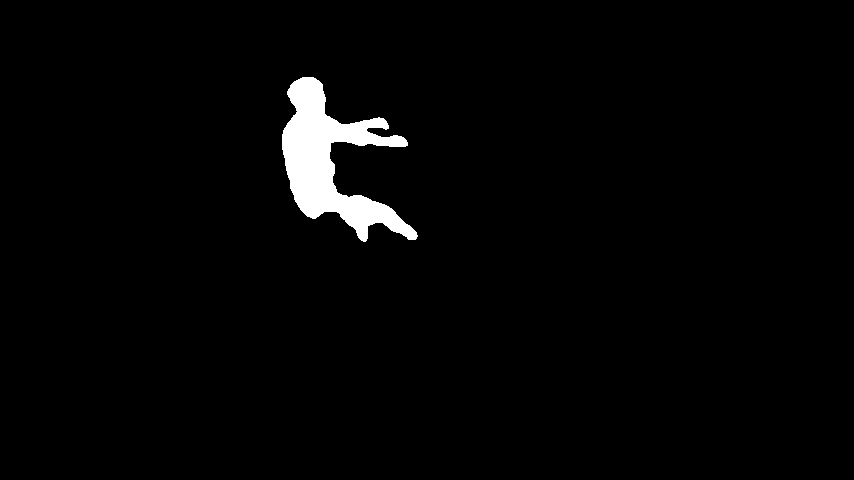}
      \caption*{\footnotesize Ours}
    \end{subfigure}
    \begin{subfigure}[t]{0.24\columnwidth}
      \centering
      \includegraphics[width=\linewidth]{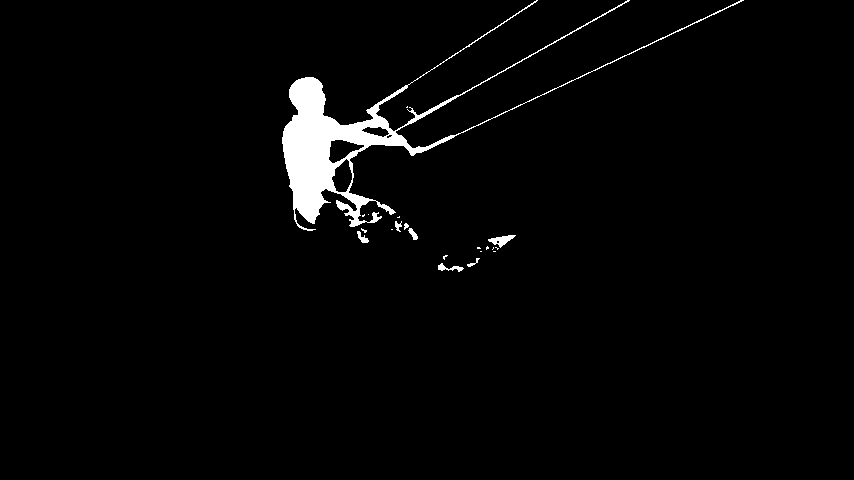}
      \caption*{\footnotesize GT}
    \end{subfigure}
  \end{subfigure}

  \caption{\textbf{Qualitative Comparison on DAVIS.} SegAnyMo yields significantly worse results when its predicted motion points fall into background regions. Our method avoids this issue by aggregating motion information within pre-segmented objects and filtering out those with low motion scores.
}
  \label{fig:davis_quali}
\end{figure}

\section{Dynamic Mask}
\noindent\textbf{Dynamic Mask Evaluation.} To demonstrate the effectiveness of our dynamic mask segmentation method, we further evaluate it on the DAVIS dataset~\cite{DAVIS2016} and compare it with recent approaches~\cite{golisabour2024romo, Huang_2025_CVPR_seganymo}. We report both \textbf{IoU} and \textbf{runtime}, averaged across all scenes. As shown in Table~\ref{tab:davis_eval}, our method achieves higher segmentation accuracy than RoMo, while running substantially faster than both SegAnyMo and RoMo. This highlights the advantage of using object-wise motion cues: our method avoids spurious motion responses in background regions and produces cleaner motion segmentation. Moreover, as illustrated in Figure~\ref{fig:davis_quali}, our approach exhibits stronger robustness than SegAnyMo, particularly in scenes where motion prediction becomes noisy or ambiguous.

We additionally provide a breakdown of the runtime in Table~\ref{tab:time_eval}. The main computational bottleneck arises from video object segmentation, especially in scenes where a large number of objects are identified by SAM~\cite{kirillov2023segany}. Nevertheless, all other components remain lightweight, and the overall runtime is still significantly lower than recent alternatives. 

\noindent\textbf{Object-wise Over-marking.}
Our object-wise masking design mitigates over-marking in two key ways:
(1) Motion averaging\ref{eq:dynmask} ensures that localized motions are diluted by the dominant static parts of an object, preventing spurious dynamic labeling.
(2) The segmentation threshold $\max(s_i)/4$ prioritizes major dynamic objects and avoids over-segmentation of minor movements.
We ablate the over-marking behavior on the \textit{Balloon2} scene. As shown in Table~\ref{tab:overmark_supp}, the over-allocation slightly increases PSNR (28.25$\to$28.45), suggesting that over-marking can be beneficial for reconstructing ambiguous scenes.

\begin{table}[h]
  \caption{Quantitative comparison of over-marking on \textit{Balloon2}.}
  \label{tab:overmark_supp}
  \centering
  \small
  \setlength{\tabcolsep}{3pt}
  \begin{tabular}{@{}lccccc@{}}
    \toprule
    Mask & PSNR & Mem (GB) & \#Static GS & \#Dynamic GS\\
    \midrule
    No Over-mark & 28.25 & 0.71 & 290{,}789 & 24{,}959 \\
    Over-mark & \textbf{28.45} & 0.66 & 261{,}218 & 35{,}152\\
    \bottomrule
  \end{tabular}
\end{table}

\noindent\textbf{Quantitative Impact of Masking Strategy.}
To isolate the contribution of our masking strategy from the Gaussian representation, we evaluate our method using different mask types. Following RoDynRF~\cite{liu2023robust}, we compute Sampson-based masks (denoted \textit{Ours-RoDynRF}). Following MoSca~\cite{Lei_mosca_2025_CVPR}, we further densify the masks using 2D tracks (denoted \textit{Ours-MoSca}). Table~\ref{tab:mask_optim_supp} demonstrates the effectiveness of our Gaussian representation.

\begin{table}[h]
  \caption{Comparison of masking strategies and optimization.}
  \label{tab:mask_optim_supp}
  \centering
  \small
  \setlength{\tabcolsep}{3pt}
  \begin{tabular}{@{}lccccc@{}}
    \toprule
     & RoDynRF & Ours-RoDynRF & MoSca & Ours-MoSca & Ours\\
    \midrule
    PSNR & 25.89 & 26.00 & 26.72 & 27.33 & \textbf{27.43} \\
    \bottomrule
  \end{tabular}
\end{table}

\section{Sensitivity Studies}

\noindent\textbf{Sensitivity study on $\beta_r$.} We conduct a sensitivity study by varying $\beta_r$ on the Nvidia dynamic scene dataset. As shown in Table~\ref{tab:s_beta_r_supp}, we vary $\beta_r$ from 1 to 10. Performance remains within $\pm$0.15\,dB of the optimum, demonstrating the robustness of our method to this threshold.

\begin{table}[h]
  \caption{Sensitivity study on $\beta_r$.}
  \label{tab:s_beta_r_supp}
  \centering
  \small
  \setlength{\tabcolsep}{3pt}
  \begin{tabular}{@{}lccccc@{}}
    \toprule
     & $\beta_r=1$ & $\beta_r=2$ & $\beta_r=4$ & $\beta_r=7$ & $\beta_r=10$\\
    \midrule
    PSNR & 27.34 & 27.43 & \textbf{27.49} & 27.46 & 27.37 \\
    \bottomrule
  \end{tabular}
\end{table}

\noindent\textbf{Robustness to Depth Prior Models.}
Since 4D reconstruction from monocular video is an ill-posed problem, it is common practice to use prior models. We mitigate errors from depth priors using a Laplacian edge filter, from optical flow using occlusion masks, and from scene flow using an intersection mask. To further demonstrate robustness, we test different depth models on the Nvidia dynamic scene dataset. As shown in Table~\ref{tab:depth_prior_supp}, the PSNR variation is small ($\pm$0.21\,dB), confirming that our method is not critically sensitive to the specific depth prior.

\begin{table}[h]
  \caption{Comparison of different depth prior models.}
  \label{tab:depth_prior_supp}
  \centering
  \small
  \setlength{\tabcolsep}{3pt}
  \begin{tabular}{@{}lcccc@{}}
    \toprule
     & Metric3D-v2 & MoGe & UniDepth & DepthPro\\
    \midrule
    PSNR & 27.43 & \textbf{27.49} & 27.28 & 27.31 \\
    \bottomrule
  \end{tabular}
\end{table}

\section{Training and Inference Comparison}
We compare training and inference costs on the DyCheck dataset in Table~\ref{tab:cost_supp}. We include results using both our default iteration count and a reduced 45K iteration setting.

\section{More Results}

We summarize the training statistics in Table~\ref{tab:stats}. We further report per-scene metrics on the DyCheck iPhone dataset in Table~\ref{tab:perscene_dycheck} for a more detailed evaluation. 

\begin{table}[h]
  \caption{Training and inference comparison on DyCheck dataset.}
  \label{tab:cost_supp}
  \centering
  \small
  \setlength{\tabcolsep}{2pt}
  \begin{tabular}{lcccc}
    \toprule
    Method & PSNR & Train.\ Time & Infer.\ FPS & Infer.\ Mem\\
    \midrule
    SoM & 17.32 & 2hrs & 144 & 1.2GB \\
    MoSca & 19.32 & 0.78hrs & 38 & 1.3GB \\
    Ours & 19.50 & 1.8hrs & 132 & 1.0GB \\
    Ours-45K & 19.44 & 0.85hrs & 196 & 1.0GB \\
    \bottomrule
  \end{tabular}
\end{table}

\begin{table}[h]
  \caption{Training Statistics}
  \label{tab:stats}
  \centering
  \resizebox{\linewidth}{!}{
  \begin{tabular}{@{}lccccc@{}}
    \toprule
    Scene & VRAM (GB) & Time (h) & \#Static GS & \#Dynamic GS & \#Total GS \\
    \midrule
    \multicolumn{6}{c}{\textbf{DyCheck iPhone Dataset}} \\
    \midrule
    Apple          & 1.68 & 1.69 & 288{,}066 & 23{,}998 & 312{,}064 \\
    Block          & 6.63 & 1.89 & 136{,}123 & 110{,}445 & 246{,}568 \\
    Paper-windmill & 0.66 & 1.65 & 387{,}559 & 24{,}845 & 412{,}404 \\
    Space-out      & 1.90 & 1.80 & 135{,}605 & 47{,}986 & 183{,}591 \\
    Spin           & 2.09 & 1.80 & 283{,}770 & 31{,}010 & 314{,}780 \\
    Teddy          & 9.12 & 2.30 & 101{,}689 & 197{,}523 & 299{,}212 \\
    Wheel          & 3.73 & 1.54 & 216{,}963 & 86{,}172 & 303{,}135 \\
    \midrule
    \textbf{Average} 
                   & 3.69 & 1.81 & 221{,}396 & 74{,}568 & 295{,}965 \\
    \midrule
    \multicolumn{6}{c}{\textbf{Nvidia Dynamic Scene Dataset}} \\
    \midrule
    Balloon1   & 3.02 & 0.29 & 259{,}834 & 66{,}593 & 326{,}427 \\
    Balloon2   & 0.71 & 0.29 & 290{,}789 & 24{,}959 & 315{,}748 \\
    Jumping    & 7.92 & 0.36 & 240{,}412 & 77{,}244 & 317{,}656 \\
    Truck      & 4.40 & 0.33 & 285{,}384 & 48{,}359 & 333{,}743 \\
    Skating    & 0.50 & 0.29 & 254{,}010 & 17{,}098 & 271{,}108 \\
    Umbrella   & 3.34 & 0.31 & 414{,}920 & 34{,}320 & 449{,}240 \\
    Playground & 1.05 & 0.31 & 370{,}305 & 30{,}399 & 400{,}704 \\
    \midrule
    \textbf{Average} 
               & 2.99 & 0.31 & 302{,}236 & 42{,}710 & 344{,}947 \\
    \bottomrule
  \end{tabular}}
\end{table}

\begin{table*}[t]
\centering
\caption{Quantitative comparison on DyCheck dataset.}
\label{tab:perscene_dycheck}
\resizebox{\textwidth}{!}{
\begin{tabular}{l|
ccc|
ccc|
ccc|
ccc}
\toprule
& \multicolumn{3}{c|}{Apple}
& \multicolumn{3}{c|}{Block}
& \multicolumn{3}{c|}{Paper-windmill}
& \multicolumn{3}{c}{Space-out}
\\
\cmidrule(lr){2-4}
\cmidrule(lr){5-7}
\cmidrule(lr){8-10}
\cmidrule(lr){11-13}
Method
& mPSNR$\uparrow$ & mSSIM$\uparrow$ & mLPIPS$\downarrow$
& mPSNR$\uparrow$ & mSSIM$\uparrow$ & mLPIPS$\downarrow$
& mPSNR$\uparrow$ & mSSIM$\uparrow$ & mLPIPS$\downarrow$
& mPSNR$\uparrow$ & mSSIM$\uparrow$ & mLPIPS$\downarrow$
\\
\midrule

Dyn. Gaussians
& 7.65 & -- & 0.766
& 7.55 & -- & 0.684
& 6.24 & -- & 0.729
& 6.79 & -- & 0.733
\\
4DGS
& 15.41 & -- & 0.456
& 11.28 & -- & 0.633
& 15.60 & -- & 0.297
& 14.60 & -- & 0.372
\\
Gaussian Marbles
& 17.70 & -- & 0.492
& 17.42 & -- & 0.384
& 17.04 & -- & 0.394
& 15.94 & -- & 0.435
\\
SoM
& 18.57 & 0.771 & 0.341
& 17.41 & 0.644 & 0.323
& 18.14 & 0.415 & 0.225
& 16.85 & 0.601 & 0.324
\\
MoSca
& 19.40 & 0.810 & 0.340
& 18.06 & 0.680 & 0.330
& 22.34 & 0.740 & 0.150
& 20.48 & 0.660 & 0.260
\\
Ours
&19.67&0.807&0.301
&18.62&0.676&0.285
&22.20&0.728&0.134
&20.43&0.680&0.243
\\
\bottomrule
\end{tabular}}

\vspace{10pt} 

\resizebox{\textwidth}{!}{
\begin{tabular}{l|
ccc|
ccc|
ccc|
ccc}
\toprule
& \multicolumn{3}{c|}{Spin}
& \multicolumn{3}{c|}{Teddy}
& \multicolumn{3}{c|}{Wheel}
& \multicolumn{3}{c}{AVE}
\\
\cmidrule(lr){2-4}
\cmidrule(lr){5-7}
\cmidrule(lr){8-10}
\cmidrule(lr){11-13}
Method
& mPSNR$\uparrow$ & mSSIM$\uparrow$ & mLPIPS$\downarrow$
& mPSNR$\uparrow$ & mSSIM$\uparrow$ & mLPIPS$\downarrow$
& mPSNR$\uparrow$ & mSSIM$\uparrow$ & mLPIPS$\downarrow$
& mPSNR$\uparrow$ & mSSIM$\uparrow$ & mLPIPS$\downarrow$
\\
\midrule

Dyn. Gaussians
& 8.08 & -- & 0.651
& 7.41 & -- & 0.690
& 7.28 & -- & 0.593
& 7.29 & -- & 0.692
\\
4DGS
& 14.42 & -- & 0.339
& 12.36 & -- & 0.466
& 11.79 & -- & 0.436
& 13.64 & -- & 0.428
\\
Gaussian Marbles
& 18.88 & -- & 0.428
& 13.95 & -- & 0.442
& 16.14 & -- & 0.351
& 16.72 & -- & 0.418
\\
SoM
& 19.35 & 0.582 & 0.247
& 13.69 & 0.542 & 0.380
& 17.21 & 0.628 & 0.230
& 17.32 & 0.598 & 0.296
\\
MoSca
& 21.31 & 0.750 & 0.190
& 15.47 & 0.620 & 0.350
& 18.17 & 0.680 & 0.230
& 19.32 & 0.706 & 0.264
\\
Ours
&20.92&0.710&0.211
&16.20&0.639&0.303
&18.45&0.692&0.205
&19.50&0.705&0.240
\\
\bottomrule
\end{tabular}}
\end{table*}
\end{document}